\documentclass{article}

\usepackage{microtype}
\usepackage{graphicx}
\usepackage{subfigure}
\usepackage{booktabs} % for professional tables

\usepackage{paralist}
\usepackage{makecell}
\usepackage{listings}
\usepackage{multirow}
\usepackage{caption}

\usepackage{hyperref}

\usepackage[T1]{fontenc}
\usepackage[accepted]{icml2025}

\usepackage{amsmath}
\usepackage{amssymb}
\usepackage{mathtools}
\usepackage{amsthm}

\newcommand{\model}{CADFusion}
\newcommand{\ttc}{Text-to-CAD}

\definecolor{codegreen}{rgb}{0,0.6,0}
\definecolor{codegray}{rgb}{0.5,0.5,0.5}
\definecolor{codepurple}{rgb}{0.58,0,0.82}
\definecolor{backcolour}{rgb}{0.95,0.95,0.92}

\lstdefinestyle{mystyle}{
    backgroundcolor=\color{backcolour},   
    commentstyle=\color{codegreen},
    keywordstyle=\color{magenta},
    numberstyle=\tiny\color{codegray},
    stringstyle=\color{codepurple},
    basicstyle=\ttfamily\footnotesize,
    breakatwhitespace=false,         
    breaklines=true,                 
    captionpos=b,                    
    keepspaces=true,                 
    numbers=left,                    
    numbersep=5pt,                  
    showspaces=false,                
    showstringspaces=false,
    showtabs=false,                  
    tabsize=4
}
\usepackage[capitalize,noabbrev]{cleveref}

\theoremstyle{plain}

\theoremstyle{definition}

\theoremstyle{remark}

\usepackage[textsize=tiny]{todonotes}

\icmltitlerunning{\ttc \ Generation Through Infusing Visual Feedback in Large Language Models}

\begin{document}

\twocolumn[
\icmltitle{\ttc \ Generation Through Infusing Visual Feedback in \\ Large Language Models}



\icmlsetsymbol{work}{$^\dagger$}

\begin{icmlauthorlist}
\icmlauthor{Ruiyu Wang}{UofT,work}
\icmlauthor{Yu Yuan}{USTC,work}
\icmlauthor{Shizhao Sun}{MSRA}
\icmlauthor{Jiang Bian}{MSRA}
\end{icmlauthorlist}

\icmlaffiliation{UofT}{University of Toronto.}
\icmlaffiliation{USTC}{University of Science and Technology of China.}
\icmlaffiliation{MSRA}{Microsoft Research Asia}

% \icmlcorrespondingauthor{Ruiyu Wang}{rwang@cs.toronto.edu}
\icmlcorrespondingauthor{Shizhao Sun}{shizsu@microsoft.com}

\icmlkeywords{Machine Learning, Computer-Aided Design}

\vskip 0.3in
]



\printAffiliationsAndNotice{{$^\dagger$ Work done during the authors' internship at Microsoft Research Asia.} 
{Open-source research project starts at March 2024.}
{Ruiyu Wang: <rwang@cs.toronto.edu>,}
{Yu Yuan: <yyhappier@mail.ustc.edu.cn>,}
% {Shizhao Sun: <shizsu@microsoft.com>, }
{Jiang Bian: <jiabia@microsoft.com>.}} % otherwise use the standard text.

\begin{abstract}
    Creating Computer-Aided Design (CAD) models requires considerable expertise and effort. \ttc, which converts textual descriptions into CAD parametric sequences, is crucial in streamlining this process. 
    Recent studies have utilized ground-truth parametric sequences, known as sequential signals, as supervision to achieve this goal.
    However, CAD models are inherently multimodal, comprising parametric sequences and corresponding rendered visual objects.
    Besides, the rendering process from parametric sequences to visual objects is many-to-one.
    Therefore, both sequential and visual signals are critical for effective training.
    In this work, we introduce \model, a framework that uses Large Language Models (LLMs) as the backbone and alternates between two training stages: the sequential learning (SL) stage and the visual feedback (VF) stage. 
    In the SL stage, we train LLMs using ground-truth parametric sequences, enabling the generation of logically coherent parametric sequences. 
    In the VF stage, we reward parametric sequences that render into visually preferred objects and penalize those that do not, allowing LLMs to learn how rendered visual objects are perceived and evaluated.
    These two stages alternate throughout the training, ensuring balanced learning and preserving benefits of both signals.
    Experiments demonstrate that \model \ improves performance, both qualitatively and quantitatively.
    Code is available at \url{https://github.com/microsoft/CADFusion}.
\end{abstract}

\section{Introduction}\label{sec:introduction}
\begin{figure*}
    \centering
    \includegraphics[width=\linewidth]{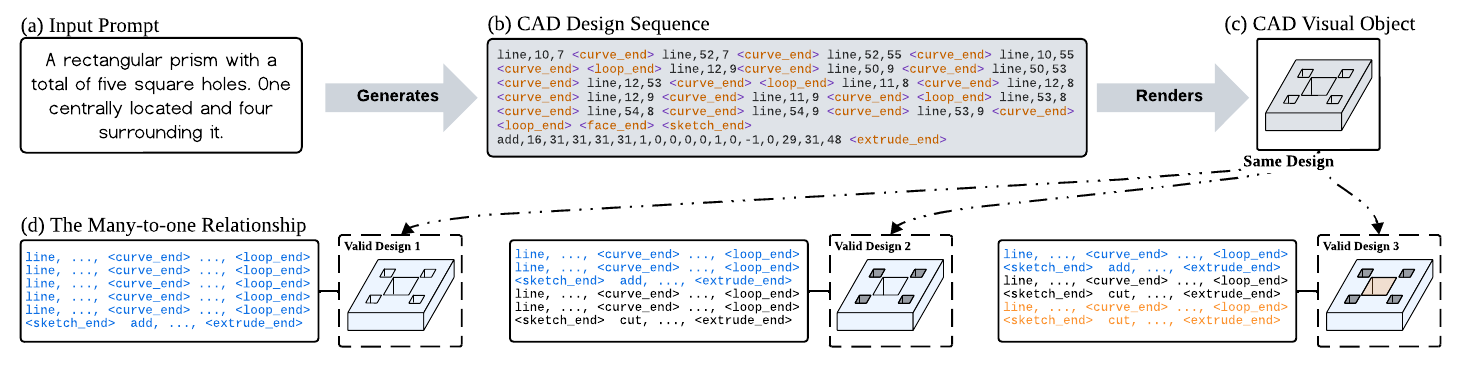}
    \vspace{-0.6cm}
    \caption{ 
    \textbf{(a) and (b):} Illustration of \ttc, which converts a textual description into CAD parametric sequences.
    \textbf{(b) and (c):} Illustration of multimodal characteristics. CAD models are created using parametric sequences and rendered as visual objects for practical use.
    \textbf{(d):} Illustration of many-to-one rendering characteristics. Different parametric sequences can produce identical visual objects.
    % The \textit{Text-to-CAD} task refer to the textual-guided \textit{(a)} CAD sequence \textit{(b)} generation.
    % CAD is multimodal as the CAD sequence generated from models \textit{(b)} is able to be rendered into visual objects \textit{(d)} for practical use.
    % Moreover, A single 3D shape \textit{(c)} can be described by multiple distinct construction designs \textit{(d)}, but learning sequentially gives only one of them the correct loss. 
    % This discrepancy causes issues where optimizing on one representation does not effectively capture the visual CAD appearance and disrupts the learning procedure.
    }
    \label{fig:representation}
\end{figure*}

Computer-Aided Design (CAD) is indispensable for 3D creation across industrial sectors~\citep{10.1115/1.4063360}.
% , 10.1016/S0360-8352(97)00138-1
It represents 3D models through a sequence of operations known as a \emph{parametric sequence}, which combines lines, arcs, and circles to create 2D sketches and then extrude them to form 3D models.
CAD models are inherently multimodal, as they are constructed using parametric sequences for precise editing and manufacturing, while also being rendered as visual objects for practical use, referred to as \emph{multimodal characteristic} (Figure~\ref{fig:representation}(b)(c)).
Moreover, the process of rendering parametric sequences into visual objects exhibits a many-to-one mapping, where different parametric sequences can result in identical visual objects, referred to as \emph{many-to-one rendering characteristic} (Figure~\ref{fig:representation}(d)).

Creating CAD models demands considerable expertise and numerous iterations, making it complex and time-consuming.
\emph{\ttc} (Figure~\ref{fig:representation}(a)(b)), which refers to the automatic generation of parametric sequences from textual descriptions, is critical for streamlining this creation process. 
It allows designers and engineers to quickly prototype and iterate designs by describing their intent in natural language, reducing the time spent on manually creating CAD models from scratch. 
Additionally, it makes the creation process more accessible to individuals without extensive training, enabling wider participation.
% It enables the generation of simple yet high-quality 3D shapes by representing objects through design sequences, such as shape sketches and 3D extrusions, and offers flexibility in the design process as shapes are not confined to unique design representations. 
% Figure \ref{fig:representation} illustrates the two core aspects that underpin the widespread adoption of this tool.
% Given that CAD design typically requires specialized training, automated CAD generation—particularly \ttc generation, which converts textual instructions into CAD designs—has emerged as a promising area of research. 
% By translating textual inputs into CAD sequences and rendering the corresponding 3D shapes, \ttc generation has the potential to significantly accelerate the design process while reducing the need for extensive pre-work training for designers.

While important, \ttc \ has received limited attention.
Most studies do not utilize text to control CAD generation.
Instead, they explore generating CAD designs from random noise~\citep{wu_deepcad_2021, xu_skexgen_2022}, by randomly altering components of existing CAD designs~\citep{xu_skexgen_2022,xu_hierarchical_2023, zhang2024flexcadunifiedversatilecontrollable}, or from point cloud~\citep{khan_cad-signet_2024}.
A few studies make preliminary attempts at \ttc~\citep{khan_text2cad_2024,li2024cad}.
They train Transformer-based framework with ground-truth parametric sequences as supervision, termed \emph{sequential signal}.

However, due to multimodal and many-to-one rendering characteristic of CAD models (Figure~\ref{fig:representation}(b)(c)(d)), both the \emph{sequential signal} and \emph{visual signal} are crucial for training a \ttc \ model.
The sequential signal, derived from ground-truth parametric sequences, provides critical information about sequence structure and parametric operations.
Without it, learning to generate logically coherent parametric sequences becomes challenging, as there is no direct supervision for sequence structure and parametric operations.
The visual signal, obtained from rendered visual objects, indicates how CAD models are perceived and evaluated in practical applications.
Without it, learning efficiency is compromised, as the goal of \ttc \ is for the rendered visual objects of the generated parametric sequences to match ground-truth visual objects.
First, sequential signal learning typically depends on auto-regressive generation, which emphasizes the local continuity between tokens but may not fully capture the global appearance of the CAD model.
Second, given the many-to-one rendering characteristic, multiple parametric sequences can produce the same visual object.
Training solely on parametric sequences may cause the model to give more emphasis to those present in the training set, overlooking other valid ones that could achieve the same visual outcome.

% Without it, learning efficiency suffers, as valid parametric sequences may be mistakenly treated as incorrect.
% This issue arises because the goal for \ttc is to generate parametric sequences that produce rendered visual objects matching the ground-truth visual objects.
% Given the many-to-one rendering characteristic, multiple parametric sequences can produce the same visual object.
% As a training set cannot to include all possible parametric sequences, a \ttc model trained solely on parametric sequences may overly prioritize those present in the training set, overlooking other valid ones that are not included but can equally achieve the desired visual object.
% This limitation hampers effective learning and reduces the model's ability to generate high-quality CAD models.

% However, due to the multimodal nature of CAD design—created as parametric sequences and rendered as visual objects for practical use—both sequential and visual signals are crucial for training a CAD model. 
% On one hand, the \textbf{sequential} signal from ground-truth CAD sequences provides essential information about sequence formatting and the geometric relationships within CAD shapes. 
% On the other hand, the \textbf{visual} signal from the rendered appearance conveys how the shapes will be perceived and assessed during actual use. 
% Neglecting the sequential signal hinders the model’s ability to capture sequence formatting, increasing invalid outputs, while ignoring the visual signal limits its ability to learn from the appearance of 3D shapes.

To this end, we propose \model, a framework that combines sequential and visual signals to train a \ttc \ model. 
It uses Large Language Models (LLMs) as its backbone and alternates between two stages: the \emph{sequential learning stage} and the \emph{visual feedback stage}.
In the sequential learning stage, LLMs are fine-tuned using ground-truth parametric sequences.
Unlike prior works~\citep{khan_text2cad_2024,li2024cad} that train Transformer-based models from scratch, we take advantage of pre-trained LLMs, which leverages their inherent natural language understanding and foundational knowledge of CAD design~\citep{makatura2023can} acquired during the extensive pre-training.
In the visual feedback stage, feedback derived from rendered visual objects is integrated into the LLMs. 
This stage addresses two critical challenges. 
First, the rendering process that converts parametric sequences into visual objects is non-differentiable, making backpropagation through this pathway infeasible. 
To overcome this, we frame the problem as preference learning task and adopt direct preference optimization (DPO)~\cite{rafailov2024direct}.
Specifically, preferences are assigned to the rendered visual objects, and the LLMs are optimized to increase the likelihood of parametric sequences that produce preferred visual objects while decreasing the likelihood of those that yield less preferred ones.
This approach enables effective training of LLMs, even with a non-differentiable rendering pathway.
Second, collecting reliable preference data is costly and labor-intensive. 
To address this, we introduce an automated pipeline that utilizes large vision-language models (LVMs) to efficiently score the rendered visual objects.
Finally, to ensure balanced learning and retain the contributions of both signals, we alternate between the sequential learning stage and the visual feedback stage throughout training.

We summarize our main contributions as follows:
\begin{compactitem}
    \item We propose to leverage both the sequential signal and visual signal to train a \ttc \ model. 
    \item For the sequential signal, we use LLMs as the backbone and fine-tune it on ground-truth parametric sequences.
    For the visual signal, we adopt direct preference optimization to bypass non-differentiable rendering and introduce a LVM-based scoring pipeline for efficient preference data collection.
    To balance both signals, we alternate between the sequential learning and the visual feedback stage.
    \item We contribute two datasets for \ttc: one with the sequential signal and another with the visual signal.
    \item We present qualitative and quantitative experiments to showcase \model’s superior ability. 
\end{compactitem}

\section{Related Works}\label{sec:related-works}

\noindent\textbf{CAD Generation.}
CAD generation takes user requirements as input and generates CAD models as output.

On the input side, user requirements can be expressed in diverse ways.
\citet{wu_deepcad_2021} uses random noise as input to generate CAD models randomly.
\citet{zhang2024flexcadunifiedversatilecontrollable}, \citet{xu_skexgen_2022} and \citet{xu_hierarchical_2023} modify specific parts of the existing CAD models to generate new ones.
\citet{khan_cad-signet_2024} and \citet{Ma_2024_CVPR} take point cloud as input to produce corresponding CAD models, while \citet{Vitruvion} uses hand sketches.
In contrast, our work focuses on textual descriptions as input, leverages LLM as backbone on stringified representations, and does not adapt codebook encoders such as VQ-VAE to generate initial representations.
Recent studies~\citep{khan_text2cad_2024, li2024cad} explore text-based input for CAD generation.
\citet{khan_text2cad_2024} proposes a data annotation pipeline for synthesizing training data and a transformer-based autoregressive network.
\citet{li2024cad} designs an encoder-decoder framework with a cascading contrastive strategy and CT-Mix to align text with parametric sequences. 
Unlike these studies, which rely solely on sequential signals, our work combines sequential and visual signals for improved performance.

On the output side, CAD models can be represented in various formats, including Constructive Solid Geometry (CSG), Boundary Representation (B-Rep) and Sketch-and-Extrude (SE).
CSG constructs 3D models by combining basic primitives such as cubes, cylinders, and spheres, through Boolean operations and subtractions \citep{10.1145/3272127.3275006, kania2020ucsgnet, yu2021caprinet, yu2023d2csg}.
B-Rep represents 3D models using geometric elements such as vertices, edges, and faces \citep{jayaraman2023solidgen, wang2022neural, xu2024brepgen}.
% 10.5555/891970, ansaldi1985geometric, 
SE begins with 2D sketches composed of lines, arcs, and circles, which are then extruded to form 3D models \citep{willis2021engineering, wu_deepcad_2021}.
In this work, we adopt SE as it preserves the design history of CAD models, making them more intuitive to edit.

\noindent\textbf{Large Language Models (LLMs).}
LLMs have recently achieved remarkable success~\citep{touvron2023llama, brown_language_2020, openai2024gpt4technicalreport, bubeck_sparks_2023, zhao_survey_2023}.
Supervised fine-tuning (SFT) is widely used to improve performance, while reinforcement learning (RL) is often employed to align LLM output with human preference~\citep{brown_language_2020, hong_knowledge_2024, kaufmann2024surveyreinforcementlearninghuman}.
Our work leverages SFT and RL\footnote{We adopt DPO~\citep{rafailov2024direct} in practice and refer to it as RL here for simplicity, as it implicitly optimizes the same objective as traditional RLHF despite not being a typical RL.} but introduces two key differences.
First, we utilize SFT and RL to learn from different signals (i.e., sequential and visual signals) whereas existing work focuses on a single signal (i.e., sequential signals).
Second, we alternate between SFT and RL stages to preserve contributions from both signals, a strategy not commonly employed in prior work.

\noindent\textbf{Reinforcement Learning with Human Feedback (RLHF).}
RLHF has been widely applied to align model output with human preference across various domains, including LLMs~\citep{brown_language_2020, radford2021learning, grattafiori2024llama3herdmodels}, text-to-image models~\citep{liang2024richhumanfeedbacktexttoimage} and text-to-video models~\citep{wu2024boosting}. 
As human annotation in RLHF is costly and not easily scalable, reinforcement learning on AI feedback (RLAIF)~\citep{liu2023gevalnlgevaluationusing, zhang2024longrewardimprovinglongcontextlarge, lee2024rlaifvsrlhfscaling}, which leverages machine learning models to annotate data, has been proposed as a more affordable alternative to RLHF.
Since RLHF/RLAIF pipeline are complex, direct preference optimization (DPO)~\citep{rafailov2024direct}, which directly optimize a model to adhere to human preferences, has been proposed to avoid explicit reward modeling or reinforcement learning.
% RLAIF leverages machine learning models to annotate data, offering a less expensive alternative to RLHF while achieving comparable performance~\citep{lee2024rlaifvsrlhfscaling}. 
In this work, we adopt DPO to address the challenge of non-differentiable rendering when learning from visual signals, as it offers a simpler yet effective solution compared to RLHF.
Besides, inspired by RLAIF, we propose an automatic scoring pipeline for CAD models using LVMs.
The generated scores are used to construct preference data, enabling efficient learning without reliance on costly human annotations.

\begin{figure*}[t]
    \centering
    \includegraphics[width=\linewidth]{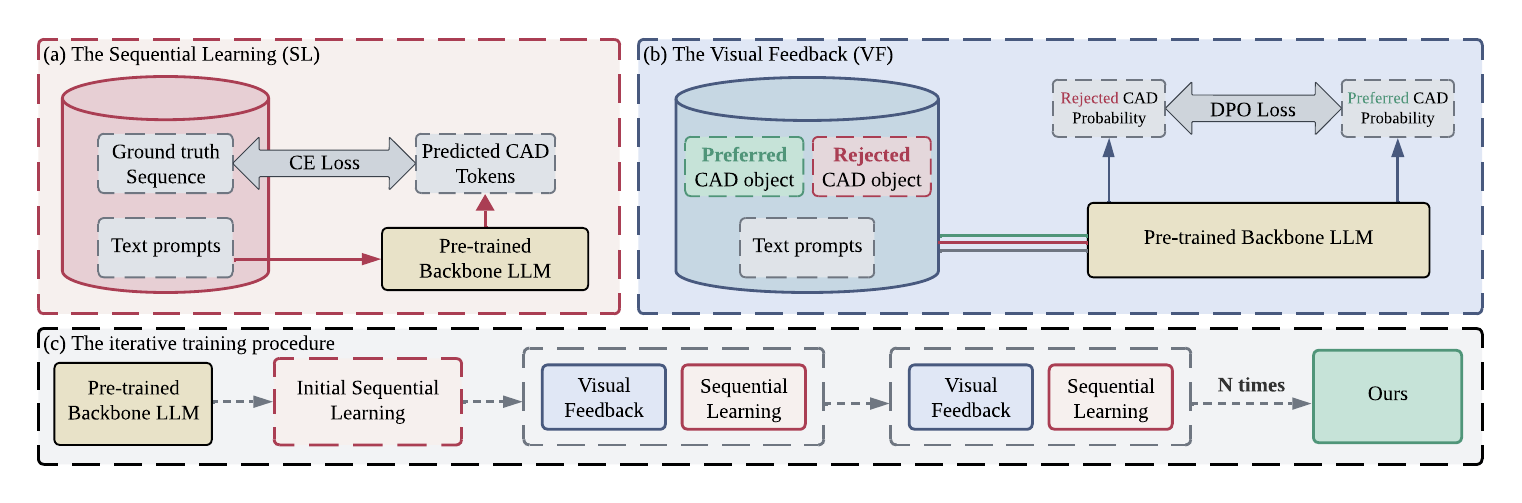}
    \vspace{-0.6cm}
    \caption{
    Overview of \model.
    \textbf{(a):} The sequential learning stage trains LLMs using ground-truth CAD parametric sequences.
    \textbf{(b):} The visual feedback stage rewards CAD parametric sequences that render into preferred visual objects and penalizes those that do not.
    \textbf{(c):} The two stages are alternated to preserve contributions of both signals.
    }
    \label{fig:pipeline}
\end{figure*}

\section{Method}\label{sec:methodology}

\subsection{Approach Overview}\label{sec:approach-overview}
Let a textual description be denoted as $x$, a CAD parametric sequence as $y$, and a rendered visual object as $o$.
The rendering process from a parametric sequence $y$ to a visual object $o$ is represented as $r(\cdot)$, such that $o=r(y)$.
\ttc \ involves learning a function $f(\cdot)$ that transforms the textual description $x$ into the CAD parametric sequence ${y}$. i.e., ${y}=f(x)$.
The goal is for the rendered visual object ${o}$ of the generated parametric sequence ${y}$, i.e., ${o}=r({y})=r(f(x))$, to match the user's desired visual object (Figure~\ref{fig:representation}).

\model \ introduces a framework that combines sequential and visual signal for training a \ttc \ model (Figure~\ref{fig:pipeline}). 
It leverages Large Language Models (LLMs) as the backbone and alternates between two stages: the \emph{sequential learning (SL) stage} and the \emph{visual feedback (VF) stage}.
We denote the model after the $i$-th round of sequential learning as $f^i_{\text{SL}}(\cdot)$ and after the $i$-th round of visual feedback as $f^i_{\text{VF}}(\cdot)$.
In the sequential learning stage, \model \ trains LLMs to learn sequence structures and parametric operations from ground-truth parametric sequences, guiding LLMs to generate logically coherent parametric sequences (Section~\ref{sec:sft}).
In the visual feedback stage, \model \ trains LLMs to understand how the rendered visual object will be perceived and evaluated. 
By rewarding parametric sequences that render into visually preferred objects and penalizing those that do not, this stage encourages LLMs to generate parametric sequences capable of producing the desired visual object (Section~\ref{sec:dpo}).
These two stages are alternated throughout training, ensuring balanced learning and preserving contributions of both signals (Section~\ref{sec:iterative}). 

\subsection{Sequential Learning Stage}\label{sec:sft}

\ttc \ requires a model capable of understanding textual descriptions and generating CAD parametric sequences that adhere to valid sequence formats and employ meaningful parametric operations. 
We adopt the following strategies to efficiently achieve these capabilities.

\noindent\emph{1) Model architecture}. We use LLMs as the backbone, leveraging their strong natural language understanding and basic CAD design knowledge~\cite{makatura2023can}.

\noindent\emph{2) CAD Parametric Sequence Format.} We adopt the format proposed by \citet{zhang2024flexcadunifiedversatilecontrollable} (Figure~\ref{fig:representation}(b)), which represents CAD parametric sequences as text tokens rather than binary representations or numerical attributes~\citep{xu_skexgen_2022,xu_hierarchical_2023,wu_deepcad_2021}. 
This text-based format simplifies processing and interpretation by LLMs.
% , including curve types (e.g., line, circle, arc), extrusion operations (e.g., add, cut, intersect), point coordinates, and extrusion parameters.

\noindent\emph{3) Dataset.} Existing CAD datasets~\citep{wu_deepcad_2021} include CAD parametric sequences but lack paired textual descriptions. 
To address this, we construct a dataset $\mathcal{D}_{\text{SL}}=\{(x,y)\}_1^M$ (`SL' for sequential learning) containing paired text $x$ and CAD parametric sequences $y$. 
We first prompt a LVM to generate draft captions for rendered CAD model images and then refine these drafts through human annotation to ensure accuracy and conciseness.
% \footnote{Details on dataset construction are provided in Appendix \ref{apdx:additional-dataset-detail}.}. 
% (e.g., GPT-4o)

\noindent\emph{4) Training.} We fine-tune the pre-trained LLMs by minimizing the discrepancy between the generated parametric sequence $\hat{y}=f_{\text{SL}}^i(x)$ and the ground-truth parametric sequence $y$ using cross entropy loss, denoted as $\mathcal{L}_{\text{SL}}$:
\begin{equation}
    \mathcal{L}_{\text{SL}} = -\mathbb{E}_{(x,y)\sim\mathcal{D}_{\text SL}}
    \left[
    \frac{1}{T}\sum_{t=1}^T\log p(\hat{y}=y_t|x)
    \right],
\end{equation} 
where $T$ is the sequence length and $p(\cdot)$ is the predicted probability of the $t$-th token by the model $f_{\text{SL}}^i(x)$.

While existing studies~\citep{khan_text2cad_2024,li2024cad} also consider sequential signals, \model \ introduces three distinctions: 1) it uses an LLM backbone to leverage pre-trained knowledge, unlike prior work that trains Transformers from scratch; 2) it represents CAD sequences as text tokens, processed with the LLM's tokenizer, whereas others use custom tokenizers; 3) its training data undergoes human annotation, while prior work relies solely on synthesized data.
These enhancements enable it to outperform existing approaches, even without the visual feedback stage. 

\begin{figure*}[t]
    \centering
    \includegraphics[width=\linewidth]{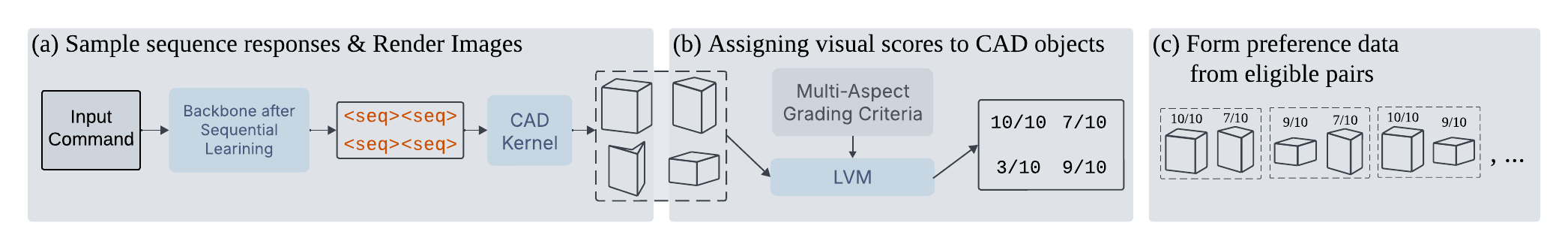}
    \vspace{-0.6cm}
    \caption{
    Illustration of preference data construction.
    \textbf{(a):} Sample CAD parametric sequences and render them into visual objects.
    \textbf{(b):} Score the visual objects using LVMs with multi-aspect grading criteria.
    \textbf{(c):} Construct preference data based on LVM-generated scores.
    }
    \label{fig:reward-modeling}
\end{figure*}

\subsection{Visual Feedback Stage}\label{sec:dpo}
The goal of \ttc \ is to ensure the rendered visual object from the generated parametric sequence matches the desired visual object. 
Relying solely on sequential signals compromises training efficiency (see Section~\ref{sec:introduction}). 
To address this, we incorporate visual feedback into the model already trained on sequential signals (i.e., $f_{\text{SL}}^i(x)$).
% The goal is that the rendered model ***. Simply learning on sequence may ***. We introduce visual signal into \model by following strategies

\noindent\textbf{Learning Visual Feedback through DPO.}
A straightforward way to incorporate visual feedback is through supervised learning, which minimizes the loss between the rendered visual object from the generated parametric sequence $\hat o=r(f(x))$, and the ground-truth visual object $o$.
However, since the rendering process $r(\cdot)$ is non-differentiable, this loss cannot be backpropagated to the model $f(\cdot)$.
To address this, we reformulate the task as a reward maximization problem, where visual feedback serves as the reward, enabling optimization without requiring a differentiable rendering process.
Since conventional RL is computationally expensive, we adopt direct preference optimization (DPO)~\citep{rafailov2024direct}, a simpler and more efficient approach that implicitly performs reward maximization.

Specifically, 
we construct a preference dataset $\mathcal{D}_{\text{VF}}=\{(x,o_w,o_l)\}_1^N$ where $o_w$ and $o_l$ are rendered from the parametric sequences $y_w$ and $y_l$, representing preferred and less preferred visual objects, respectively.
We then optimize the model to increase the likelihood of parametric sequences that produce preferred visual objects ($y_w$), while decreasing the likelihood of those that yield less preferred ones ($y_l$):
\begin{align}
        \mathcal{L}_{\text{VF}} & =  -\mathbb{E}_{(x,y_w,y_l)\sim\mathcal{D}_{\text{VF}}} \\ \nonumber
        & \left[
        \text{log}\sigma(
            \beta \log \frac{p(\hat y=y_w|x)}{p_{\text{ref}}(\hat y=y_w|x)} - 
            \beta \log \frac{p(\hat y=y_l|x)}{p_{\text{ref}}(\hat y=y_l|x)}
        ) 
    \right],
\end{align}
where $p(\cdot)$ is the predicted probability of a parametric sequence under the current model ($f_{\text{VF}}^i(x)$), $p_{\text{ref}}(\cdot)$ the probability under the reference model from the last round of sequential learning ($f_{\text{SL}}^i(x)$) and $\beta$ is scaling factor.

\noindent\textbf{Constructing Preference Data with LVM Scoring.}
Collecting preference data is both costly and labor-intensive. 
The iterative use of the visual feedback stage in our framework (Section~\ref{sec:iterative}) further highlights the need for a quick and efficient approach for obtaining preference data. 
To address this, we propose leveraging the strong visual understanding capabilities of LVMs to score visual objects and construct preference data.
Figure~\ref{fig:reward-modeling} outlines the pipeline. 
First, the textual description $x$ is input into the finetuned model after sequential learning ($f_{\text{SL}}^i$) to generate multiple parametric sequences, which are then rendered into visual objects (e.g., CAD images in our implementation). 
Next, the rendered CAD images, along with an instruction detailing the evaluation criteria, are input into an LVM to obtain scores. 
% (e.g., LLaVA-OneVision~\citep{li2024llava} in our implementation)
Finally, the CAD image with the higher score is regarded as the preferred one (i.e., $o_w$), while the one with the lower score is deemed as the less preferred one (i.e., $o_l$).

\begin{figure}
    \centering
    \includegraphics[width=\linewidth]{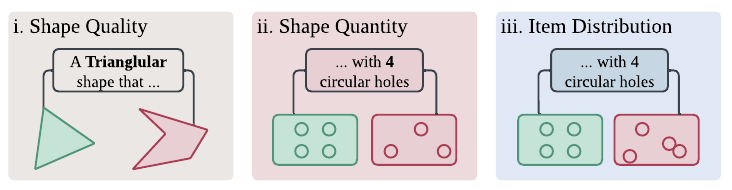}
    \vspace{-0.6cm}
    \caption{
    An illustrative example of the multi-aspect evaluation criteria used in LVM scoring.
    Note that the illustrations are simplified to conceptually represent each criterion.
    }
    \label{fig:reward-criteria}
\end{figure}

Specifically, inspired by recent work~\citep{liang2024richhumanfeedbacktexttoimage} on evaluating text-to-image generation across rich aspects, we incorporate multiple evaluation criteria into the LVM instruction. 
As shown in Figure~\ref{fig:reward-criteria}, these criteria assess both the appearance of CAD designs and their alignment with textual descriptions: 1) \emph{shape quality} evaluates the regularity, naturalness, and realism of the design; 2) \emph{shape quantity} checks whether the number of components matches the description; and 3) \emph{distribution} ensures components are arranged naturally, avoiding collisions or excessive spacing.

\subsection{Alternate Training}\label{sec:iterative}
Each stage of the training process — sequential learning and visual feedback — has a specialized focus. 
Excessive training in one stage can lead to the degradation of skills acquired in the other.
For example, we empirically observe that extended training with visual feedback can impair the model's ability to generate well-formatted parametric sequences, a skill developed during sequential learning. 
Conversely, prolonged training with sequential signals can weaken the model's capacity to produce parametric sequences that render visually natural objects, a capability enhanced during the visual feedback stage.
To mitigate this, we introduce an alternate training strategy (Figure~\ref{fig:pipeline}(c)). 
The process begins with the sequential learning stage, ensuring the model acquires the ability to generate logically coherent parametric sequences. 
Subsequently, the training is divided into smaller blocks. 
Within each block, the model first learns from the visual signal, followed by the sequential signal, balancing the two objectives effectively.
% Within each block, the model alternates between receiving visual signals and sequential signals, balancing the two objectives effectively.

\section{Experiments}\label{sec:experiments}
\subsection{Setups}\label{sec:implementation}
\noindent\textbf{Datasets.}
% Sequential learning: how (a very brief intro + refer to the main body/appendix) & how many data items
% Visual feedback: how (a very brief intro + refer to the main body/appendix) & how many data items & which LVM is used
% Test split
For the dataset used in the sequential learning stage, we use DeepCAD dataset~\citep{wu_deepcad_2021} as the source for CAD parametric sequences (specifically the version processed by \citet{xu_skexgen_2022}).
We construct a dataset compromising 20k pairs of textual instructions and CAD parametric sequence using the techniques introduced in Section~\ref{sec:sft} and Appendix~\ref{apdx:dataset-construction}.
For the preference data used in the visual feedback stage, we employ \texttt{llava-onevision-qwen2-7b}~\citep{li2024llava} to construct it using the method introduced in Section~\ref{sec:dpo}.
For each iteration of the visual feedback, we generate approximately 1,500 preference pairs, by using 1,000 text prompts as input, sampling 5 times per prompt, and filtering out invalid or low-quality samples.
For the test set, we construct it by splitting the dataset used in sequential learning into train, validation, and test sets with a 90:5:5 ratio.
% by sampling 1,000 prompts five times and filtering out invalid and low-quality samples.

% We utilize the SkexGen dataset \citep{xu_skexgen_2022}, which contains approximately 170k CAD sequences. 
% From this dataset, we extract around 20k data instances using the method described in Section \ref{apdx:dataset-construction}.
% For visual learning, we employ \texttt{llava-onevision-qwen2-7b-ov-chat} as the LVM judge. 
% In each iteration, we generate approximately 1,500 preference pairs by sampling 1,000 prompts five times and filtering out invalid and low-quality samples.
% The sequential learning dataset is split into training, validation, and test sets with a 90:5:5 ratio.
\noindent\textbf{Implementation Details.}
% Backbone
% Hyperparameters in the first round sequential learning stage: LoRA, epoch, lr, optimizer, etc.
% Hyperparameters in the visual feedback stage & other sequential learning stage: LoRA, epoch, lr, optimizer, etc.
% how many iterations of visual+sequential 
% How many GPUs (time is not necessary)
\texttt{LLaMA-3-8b-Instruct} is used as the LLM backbone, with a maximum token length of 1024. 
For efficient fine-tuning, we adopt Low-Rank Adaptation (LoRA) \citep{hu2021loralowrankadaptationlarge} with hyperparameters $r=32$ and $\alpha=32$.
The initial sequential learning stage lasts for 40 epochs with a learning rate of $1 \times 10^{-4}$, using the AdamW optimizer. 
Following this, we run 5 iterations of alternating visual feedback and sequential learning stages. 
The visual feedback stage lasts for 5 epochs on the preference data, while the sequential learning stage lasts for 1 epoch using the same dataset as the initial sequential learning stage.
Training is conducted on four NVIDIA A6000-48GB SMX GPUs using PyTorch Distributed Data Parallel (DDP).
% For the initial sequential learning stage, we apply Low-Rank Adaptation (LoRA) \citep{hu2021loralowrankadaptationlarge} with hyperparameters $r=32,~\alpha=32$, training for 40 epochs with an initial learning rate of $1\times10^{-4}$ using the AdamW optimizer.
% During the visual feedback stage, sequences are sampled with a temperature of 0.9 and top\_p = 0.9. 
% Each visual learning iteration consists of performing DPO on the generated data for 5 epochs, followed by 1 epoch of sequential learning. 
% This iterative process is repeated five times.
% Training is conducted on four NVIDIA A6000-48GB SMX GPUs using PyTorch Distributed Data Parallel (DDP).

\noindent\textbf{Baselines.}
% Text2CAD
% GPT-4o
We consider two types of baselines.
The first is a specialized model for \ttc~\citep{khan_text2cad_2024,li2024cad}. 
We use \citet{khan_text2cad_2024} as our baseline, as \citet{li2024cad} is not open-sourced and we were unable to reproduce it ourselves.
The second baseline is a general model that acquires some CAD knowledge during pre-training. 
We use the most powerful model, GPT-4o, as our baseline. 
Specifically, we apply few-shot learning, providing 8 examples as input for GPT-4o.
% For baseline comparisons, we evaluate an LLM-based few-shot method using \texttt{GPT-4o}, in which we provide an 8-shot example randomly sampled from the training set.
% Additionally, we use the checkpoint released by \citet{khan_text2cad_2024} to assess the performance of Text2CAD. 
% Since our prompts do not include specific location coordinates or precise geometric details, we categorize them as intermediate-level (L2) instructions following the classification in \citet{khan_text2cad_2024}. 
% A detailed justification is provided in Section \ref{apdx:text2cad}.

\noindent\textbf{Metrics.}
% sequence-level metrics
% visual-level metrics
% human evaluation
Our evaluation focuses on assessing the alignment of generated CAD models with input instructions and the overall quality of the generated CAD models. 
% We employ the following metrics to evaluate both sequential and visual level quality in model generation.
We employ the metrics at both the sequential level and visual level.
First, to evaluate the correspondence between the ground-truth and generated parametric sequences, we use F1 scores following \citet{khan_text2cad_2024}. 
Specifically, we compute F1 score for primitives (averaged over lines, arcs, and circles for brevity) and extrusions, denoted as \textbf{F1-Sketch} and \textbf{F1-Extrusion}.
% Following \citet{khan_text2cad_2024}, we measure the F1 scores of primitives and extrusions, but for brevity we average the score of primitives (lines, arcs, and circles) and call it sketch score.
Second, to assess the quality of the generated CAD models, we compare the ground-truth and generated point clouds. 
We adopt Chamfer Distance (\textbf{CD}) from \citet{khan_text2cad_2024} and additional metrics from \citet{xu_skexgen_2022}, including Coverage (\textbf{COV}), which quantifies the percentage of real data covered by generated samples using CD; Minimum Matching Distance (\textbf{MMD}), which evaluates the closest match between generated samples and real data; and Jensen-Shannon Divergence (\textbf{JSD}), which measures distribution similarity.
Additionally, we compute the Invalidity Ratio (\textbf{IR}), which quantifies the percentage of generated parametric sequences that fail to render into valid visual objects.
Furthermore, we introduce an LVM-based metric, denoted as \textbf{LVM Score}, to assess the visual correspondence between model predictions and input instructions. 
To this end, we employ \texttt{GPT-4o} with a dedicated evaluation prompt. Further details are provided in Appendix \ref{apdx:additional-exp-setup}.
Finally, we conduct human assessments to rank generations from different baselines, denoted as \textbf{Avg. Rank}. Details on this evaluation can be found in Appendix \ref{apdx:human-eval}.

\begin{table*}[t]
    \small
    \centering
    \begin{tabular}{lccccccccc}
    \Xhline{2\arrayrulewidth}
    \multirow{2}{*}{\textbf{Method}} & \multicolumn{2}{c}{\textbf{F1}$\uparrow$} & \multirow{2}{*}{\textbf{CD$\downarrow$}} & \multirow{2}{*}{\textbf{COV} $\uparrow$} & \multirow{2}{*}{\textbf{MMD} $\downarrow$} & \multirow{2}{*}{\textbf{JSD} $\downarrow$} & \multirow{2}{*}{\textbf{IR} $\downarrow$} & \multirow{2}{*}{\textbf{LVM Score} $\uparrow$} & \multirow{2}{*}{\textbf{Avg. Rank} $\downarrow$}\\
    & Sketch & Extrusion & \\
    \midrule
        % \textbf{GPT-4o} \textit{-3shot} (*) & 54.67 & 12.61 & 61.24 & 92.01 & - & - \\
        % \textbf{\model}-\textit{w/o iterative SFT} (*) & 74.00 & 6.75 & 24.88 & 88.87  & - & - \\ 
        % \textbf{\model}-\textit{SFT only} & 72.80 & 3.57 & 12.22 & \textbf{5.56} & 7.69 & 2.03 \\  
        % \textbf{Text2CAD} & 1 & 2 & 3 & 72.39 & 3.60 & 11.56 & 20.90  & 6.87 & 2.97 \\
        % \textbf{Text2CAD} & 24.59 & 75.37 & 235.91 & - & - & - & 3.37  & 2.01 & 2.97 \\ % our test set, our prompt. Also, invalidity = 24.16 according to our renderer
        % \textbf{Text2CAD3} & 39.40 & 93.56 & 146.15 & - & - & - & 3.91  & 2.01 & 2.97 \\ % our test set, t2c intermediate prompt 
        \textbf{GPT-4o} & 82.96 & 85.72 & 68.50 & 72.40 & 6.60 & 37.93 & 74.26  & 5.13 & 3.22 \\ 
        \textbf{Text2CAD} & 63.94 & {92.13} & 30.23 & - & - & - & \textbf{3.37} & 2.01 & 2.97 \\ % our test set, t2c expert prompt
        \textbf{\model} & \textbf{85.22} & \textbf{92.79} & \textbf{19.89} & \textbf{90.40} & \textbf{3.49} &\textbf{17.11} & 6.20 & \textbf{8.96} & \textbf{1.86} \\  
        \Xhline{2\arrayrulewidth}
    \end{tabular}
    \caption{Quantative results - Test results on F1 scores including \textbf{Sketch} (primitive, averaged) and \textbf{Extrusion}, Chamfer Distance (\textbf{CD}), Coverage (\textbf{COV}), Minimum Matching Distance (\textbf{MMD}), Jensen-Shannon Divergence (\textbf{JSD}), Invalidity Ratio (\textbf{IR}), the \textbf{LVM Score} and the average rank from human evaluation (\textbf{Avg. Rank}). An upward arrow ($\uparrow$) indicates that higher values are better, while a downward arrow ($\downarrow$) signifies that lower values are preferred. 
    Since Text2CAD does not release COV, MMD, and JSD, and we were unable to compute them ourselves due to differences in setup, these values are unavailable.
    % Due to differences in setup, Text2CAD results are not reported for COV, MMD, and JSD. 
    % For the other two models, these metrics are evaluated using 50 samples over five runs.
    }
    \label{tab:quantative}
\end{table*}

\begin{figure*}[t!]
    \centering
    \includegraphics[width=\linewidth]{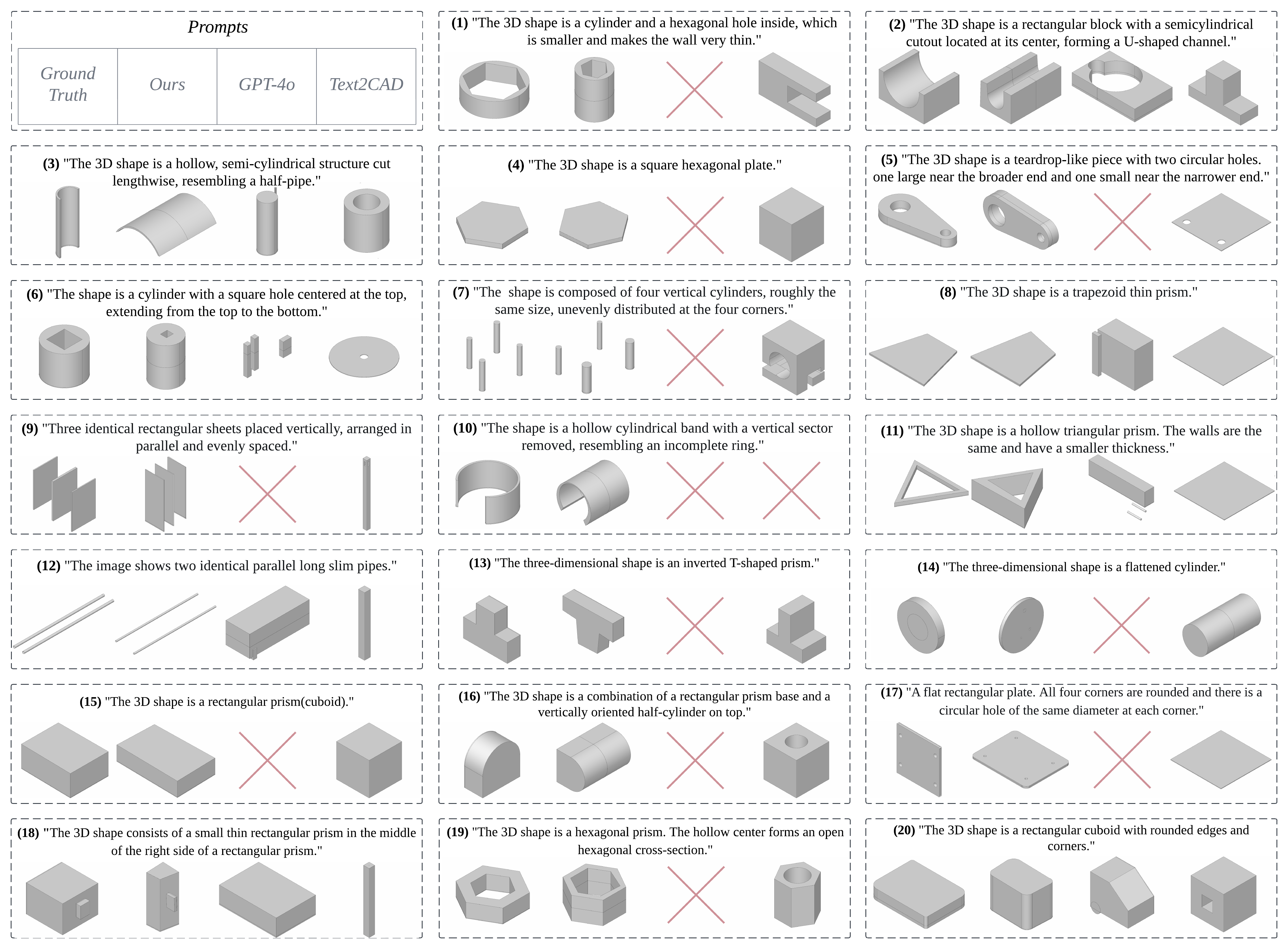}
    \caption{
    Qualitative results.
    The input prompt is shown at the top of each subsection. 
    Images are arranged from left to right in the following order: ground truth, \model, GPT-4o, and Text2CAD. 
    Outputs that cannot be rendered are marked with a red cross.
    \model\ outperforms all baselines in understanding instructions and generating CAD objects that are both sequentially and visually high quality. 
    GPT-4o frequently produces invalid samples and pays little attention to shape details. 
    Text2CAD generates well-formed basic shapes with a regular appearance but struggles to accurately follow input instructions and represent complex geometries.
    % Qualitative results - The input prompt is shown at the top of each subsection. Images are arranged from left to right as follows: ground truth figure, \model\ generation, GPT-4o results, and Text2CAD generations. Outputs that failed to render are marked with a red cross.  GPT-4o generates mostly invalid samples with limited attention to shape details. 
    % % comment out the line below if insufficient space
    % Text2CAD produces high-quality basic shapes but performs poorly in responding to prompt instructions. Our model outperforms both baselines in understanding instructions and generating sequentially and visually high-quality CAD objects.
    }
    \label{fig:qualitative}
\end{figure*} 

\subsection{Main Results}\label{sec:experimental-results}

\noindent\textbf{Quantitative Evaluation.}
Table \ref{tab:quantative} summarizes the quantitative results comparing \model\ with baseline methods (see Appendix \ref{apdx:text2cad} for more details). 
Compared to \texttt{GPT-4o}, \model\ outperforms it across all metrics.
This suggests that while the general model may have acquired some CAD knowledge during pre-training, explicitly optimizing for \ttc, as in our approach, is crucial for improving performance, 
Compared to Text2CAD\footnote{Our comparison with Text2CAD is not entirely aligned and is in favor of it.
Performing poorly on prompts we provided, we have present the results of Text2CAD tested with their original prompts.
We detail this problem in Appendix \ref{apdx:text2cad}.}, \model\ achieves comparable or better performance on all metrics, with particular strengths in metrics reflecting the visual quality such as LVM score and Avg. Rank.
This highlights the effectiveness of incorporating visual signals in our approach, as opposed to Text2CAD, which relies solely on sequential signals.
This outcome also aligns with \citet{khan_text2cad_2024}'s limitation statement that Text2CAD is limited to generating only rectangular and cylindrical shapes. 
When faced with complex geometries, it struggles to perform effectively.

% Our model outperforms \texttt{GPT-4o} in all aspects, including both sequence quality and visual quality.
% Compared to Text2CAD, \model\ achieves the highest F1 scores on sketches and delivers comparable performance on extrusions, CD, and IR. 
% Due to differences in experimental setups, we cannot fully align Text2CAD outputs with metrics not originally reported in their work. 
% Additional details on discrepancies between ours and the Text2CAD setup are provided in Appendix \ref{apdx:text2cad}.

% In visual evaluations, \model\ is ranked highest by human judges and achieves the highest LVM score of 8.96. 
% In contrast, Text2CAD receives lower rankings in these criteria, and a low score. 
% This outcome aligns with \citet{khan_text2cad_2024}'s limitation statement that Text2CAD is limited to generating only rectangular and cylindrical shapes. 
% When faced with complex geometries, it struggles to perform effectively. 
% This limitation is reflected not only in its LVM score but also in its low F1 score on arcs, which reduces the average sketch score, as well as in qualitative assessments.

\noindent\textbf{Qualitative Evaluation.}
Figure \ref{fig:qualitative} compares the results among the ground truth, our method, GPT-4o, and Text2CAD on the test set.
% As shown in Figure \ref{fig:qualitative}, 
GPT-4o frequently fails to produce renderable results across most test cases, which aligns with its high invalidity ratio (IR) reported in Table \ref{tab:quantative}. 
% Despite sometimes generating similar shapes, its responses in most cases are poorly aligned with the input prompts. 
While it occasionally generates valid shapes, its outputs are often misaligned with the input prompts.
% Text2CAD demonstrates high output quality, generating well-formed shapes without irregular edges or corners. 
Text2CAD generate well-formed shapes without irregular edges or corners. 
However, it often produces oversimplified shapes and, for more complex prompts, tends to generate multiple cubes or panels instead of accurately capturing the intended structure.
This aligns with its low invalidity ratio (IR) but poor visual scores. such as LVM score and Avg. Rank, in Table \ref{tab:quantative}.
\model\ provides the most precise response to input instructions and achieves the highest similarity to the ground truth. 
It successfully captures complex shapes, including rectangles, hexagons, and nested structures, such as a hexagonal hole within a cylinder. 
Additionally, it exhibits a strong understanding of language cues, accurately interpreting numerical and qualitative descriptors like ``long" or ``T-shape".
Additional qualitative results, as well as our model’s ability to generate multiple varied outputs, are presented and discussed in Appendix \ref{apdx:qualitative} and \ref{apdx:text-to-multiple-cad}.

\begin{table}[t]
    \centering
    \begin{tabular}{lcc}
    \Xhline{2\arrayrulewidth}
    \textbf{Method} & \textbf{LVM Score} $\uparrow$ & \textbf{IR} $\downarrow$ \\
    \midrule
    % invalidity will be update again
         \model$_{\text{SL}}$ & 7.69 & 4.84\\
         \model$_{\text{SL}_{\text{w/o HA}}}$ & 6.56 & 6.00 \\
         \model$_{\text{SL-VF}}$ & 5.94 & 88.87 \\
         \model$_{\text{SL-VF}_{\text{RPO}}}$ & 6.21 & 3.46\\
         \model$_{\text{SL-VFSL(1)}_{\text{w/ HA}}}$ & 8.28 & 17.03\\
         \model$_{\text{SL-VFSL(1)}}$ & 8.76 & 4.42\\
         \model$_{\text{SL-VFSL(3)}}$ & 8.89 & 4.21\\
         \model$_{\text{SL-VFSL(5)}}$ & 8.96 & 6.20\\    
     \Xhline{2\arrayrulewidth}
    \end{tabular}
    \caption{LVM scores and invalidity ratios across different \model\ variants. The suffix \texttt{SL} indicates that the model is trained with the initial Sequential Learning stage, while \texttt{VF} denotes the Visual Feedback stage without additional Sequential Learning. \texttt{VFSL} represents Visual Feedback with alternating Sequential Learning. The tag \texttt{w/ HA} signifies that the data is preprocessed with human annotation, whereas \texttt{w/o HA} denotes the absence of human annotation. Numbers in parentheses indicate the number of \texttt {VFSL} rounds performed. \texttt{RPO} refers to the model using Regularized Preference Optimization (RPO) \citep{liu2024provablymitigatingoveroptimizationrlhf} to stabilize DPO.
    % instead of our proposed method.
    }
    \label{tab:lvm-scores}
\end{table}

\subsection{Ablation Studies} \label{sec:ablation}
We conduct ablation studies on the effectiveness of the visual feedback stage, the impact of the alternate training, and the choice between human and LVM annotation for data.

\noindent\textbf{Visual Feedback.}
To assess the importance of visual feedback, we conduct an ablation study on \model\ using only sequential learning, denoted as \model$_{\text{SL}}$. 
The first row of Table \ref{tab:lvm-scores} presents its LVM score and invalidity ratio. 
Compared to our approach, denoted as \model$_{\text{SL-VFSL(5)}}$ in Table \ref{tab:lvm-scores}, while \model$_{\text{SL}}$ improves the invalidity ratio by 1.36\%, it results in a considerable decrease in the LVM score. 
% This underscores the crucial role of the visual feedback stage: by leveraging visual preference data, our framework effectively enhances the model’s visual generation quality.
This underscores the crucial role of the visual feedback stage: by leveraging visual preference data, our framework effectively enhances the visual quality of the generated CAD models.
Additionally, \model$_{\text{SL}}$ outperforms the baseline method, Text2CAD, which also relies solely on sequential signals. 
Note that this advantage is achieved using 20k data, while Text2CAD uses 150k data.
This demonstrates the effectiveness of the techniques employed in our sequential learning stage, including leveraging LLMs as the backbone, representing CAD parametric sequences as textual tokens, and utilizing human annotations (Section~\ref{sec:sft}).
% should be better than text2cad to justify that the tricks in our sequential learning stage indeed work.

\noindent\textbf{Alternate Training.}
In Section \ref{sec:iterative}, we propose an alternate training strategy to retain the benefits of both sequential learning and visual feedback stage.
We compare this approach with three variations:
1) visual feedback only (\model$_{\text{SL-VF}}$), 2) visual feedback with an additional Negative Log Likelihood loss (\model$_{\text{SL-VF}_{\text{RPO}}}$) to regularize and stabilize DPO~\citep{liu2024provablymitigatingoveroptimizationrlhf}, and 3) iterative visual-sequential training (our method).

% In Section \ref{sec:iterative}, we hypothesize that an alternate training procedure would mitigate DPO’s instability, which impairs \model's ability to maintain formatting consistency. 
% After the same sequential learning round, we compare three training variations: (1) Visual feedback only (\model$_{\text{SL-VF}}$), (2) Visual feedback with an additional Negative Log Likelihood loss (\model$_{\text{SL-VF}{RPO}}$)\footnote{Regularized Preference Optimization (RPO) \citep{liu2024provablymitigatingoveroptimizationrlhf}, a common method for regularizing and stabilizing DPO.}, and (3) Iterative visual-sequential training (our method).

Table \ref{tab:lvm-scores} presents the results, with our approach denoted as \model$_{\text{SL-VFSL(5)}}$.
The high invalidity ratio of \model$_{\text{SL-VF}}$ indicates that it struggles to generate renderable sequences, suggesting that extended training with visual signals can impair the model's ability to generate well-formatted parametric sequences.
Besides, \model$_{\text{SL-VF}}$ receives a low rating from the LVM judge, revealing that training with visual feedback along provides limited benefit.
Regarding \model$_{\text{SL-VF}_{\text{RPO}}}$ which incorporates the additional loss, while it achieves low invalidity ratio, its visual quality, as assessed by the LVM judge, is even lower than the SL-only setup (i.e., \model$_{\text{SL}}$).
This indicates that it fails to effectively balance the contributions of both sequential signals and visual signals.

% The results, shown in Table \ref{tab:lvm-scores}, reveal that training with visual feedback alone provides no benefit to the framework, as \model$_{\text{SL-VF}}$ is rated poorly by the LVM judge. 
% Moreover, the high invalidity ratio indicates that it struggles to generate renderable sequences, reinforcing our hypothesis regarding DPO’s instability in our setup.
% Regarding \model$_{\text{SL-VF}_{RPO}}$ which adopts the additional loss, while it achieves low invalidity ratio, its visual quality, as assessed by the LVM judge, is even lower than the SL-only setup, raising questions about its overall effectiveness. 
% Our method, while maintaining a similar invalidity level, significantly improves visual quality compared to \model\ after sequential learning and all previously discussed variants.

% We also analyze results across multiple training iterations. 

We also compare model variants that use different numbers of iterations of visual feedback and sequential learning.
In Table \ref{tab:lvm-scores}, for each \model$_{\text{SL-VFSL(*)}}$ variant, the number in parentheses indicates the number of alternative training rounds performed. 
The results for iterations 1, 3, and 5 are reported, showing a gradual increase in LVM scores along with a stable invalidity ratio. 
This further validates the effectiveness of our approach.

\noindent\textbf{Data Annotation.}
We examine the impact of our choice of data annotation. 
In the sequential learning stage, the dataset is constructed by first using LVMs to generate initial captions, followed by human annotators refining them. 
To evaluate the effect of this decision, we conduct an experiment in which our method is trained on data without human annotation, denoted as \model$_{\text{SL}_{\text{w/o HA}}}$. 
The second row of Table \ref{tab:lvm-scores} presents the results. 
It shows worse LVM score and IR compared to the version using data with human annotations (\model$_{\text{SL}}$), highlighting the necessity of human annotation in the sequential learning stage.

% Here, we examine our choice of data annotation. 
% Human annotation is incorporated to enhance our \ttc\ data for sequential learning after LVM captioning. 
% To evaluate the impact of this decision, we conduct an ablation study in which our framework is trained on data without human involvement, denoted as \model$_{\text{SL}{\text{w/o HA}}}$ (SL without Human Annotation). 
% The second row of Table \ref{tab:lvm-scores} presents the results. 
% According to the LVM evaluation, our chosen sequential learning setup achieves a visual quality score more than one point higher than its non-human-annotated counterpart, demonstrating a clear advantage of human annotation.

In the visual feedback stage, LVMs are used to score CAD models and generate preference data.
This design choice is driven by the high cost of human annotation and the challenge of managing human annotators to ensure consistent scoring.
To evaluate the effect of this decision, we conduct an experiment where the visual feedback stage of our method is trained on human-scored preference pairs, denoted as \model$_{\text{SL-VFSL(1)}_{\text{w/ HA}}}$.
Compared to the LVM-scored version (i.e., \model$_{\text{SL-VFSL(1)}}$), it achieves a worse LVM score and IR.
This aligns with our intuition that, while human annotation may be more accurate, managing annotators for consistent scoring is difficult.
Furthermore, using LVM-scored preference data allows \model$_{\text{SL-VFSL(1)}}$ to scale across more rounds of visual feedback (e.g., \model$_{\text{SL-VFSL(5)}}$), leading to improved performance.
Achieving this with human annotation would be challenging and expensive.

% Additionally, we use LVM scoring during the feedback stage for scalability.
% \model$_{\text{SL-VFSL(1)}{\text{w/ HA}}}$ denotes the variant trained on human-scored preference pairs. 
% Comparing its results to the standard VFSL model at the same epoch (\model$_{\text{SL-VFSL(1)}}$) reveals that human annotation leads to slightly lower LVM scores and a significant increase in invalidity.
% We speculate that this may be due to LVM judges exhibiting stronger instruction-following capabilities. 
% Given the multi-aspect evaluation criteria, LVM judges may provide more informative feedback than human evaluators, who are only instructed to rank images based on quality.
% HA = Humman Annoatation

\section{Limitation}
\model's results are overall promising. 
However, there are limitations that could be addressed in future work.
First, modern LVMs suffer from performance drop when handling multiple images as input. 
Currently, we can only provide LVM with a single-view image to ensure both accurate image understanding and prompt following. 
This limitation prevents us from achieving a more effective Visual Feedback pipeline and evaluator.
Second, \model\ struggles to generate very complex shapes that require spatial and commonsense reasoning, such as the shapes of letters and words (see Appendix \ref{apdx:failure}). 
% We anticipate that a more advanced backbone will help address these challenges.

\section{Conclusion}\label{sec:conclusion}
We propose \model\ for \ttc, the first approach to incorporate visual feedback from rendered CAD objects into the training pipeline. 
\model\ uses LLMs as backbone and alternates between the sequential learning stage and the visual feedback stage.
We conduct extensive experiments to demonstrate the superiority of \model\ and validate the effectiveness of the design choices.
In the future, we plan to further improve the preference data construction pipeline to enhance performance, and collect more CAD data with more complex geometric shapes to investigate \model’s performance on more challenging cases.
% In this work, we propose \model, an LLM-based framework for \ttc\ generation. 
% To the best of our knowledge, this is the first approach that incorporates visual quality into the training pipeline. 
% Our method alternates between training LLMs with sequential learning and refining them through visual feedback. 
% In sequential learning, LLMs are trained on ground-truth CAD sequences, while in the visual feedback stage, the backbone is fine-tuned using preference pairs scored based on rendered visual appearances.
% Through extensive qualitative and quantitative evaluations, we demonstrate \model's superiority over existing baselines in both sequential and visual quality. 
% Our in-depth ablation studies highlight the benefits of training with visual information and validate key design choices, including our alternate training paradigm.
% By enhancing \ttc\ generation quality, we envision that our method will accelerate the CAD design process while reducing the need for intensive training and specialized expertise.

\section*{Acknowledgement}
We thank Weijian Ma for insightful conversation and discussion. 
We also appreciate the informative comments and suggestions provided by the anonymous reviewers.

\section*{Impact Statement}
This paper presents work aimed at improving \ttc \ generation through the use of LLM-based frameworks and the incorporation of visual feedback. 
Our work has the potential to enhance the CAD design process, offering the benefits of automation and efficiency while reducing reliance on intensive training and specialized expertise. 
This could make CAD design more accessible, particularly in industries where skilled designers are in short supply or where rapid prototyping is essential.

\section*{Ethics Statement}
In this work, we have invited crowd workers to give textual descriptions to CAD models. 
We conducted this work in accordance with ethical guidelines to ensure that participants were treated fairly, respectfully, and safely throughout the process. 
We took steps to protect the privacy of crowd workers by not collecting personally identifiable information. 
The data annotated by the crowd workers was used only for research purpose related to improving CAD generating techniques.

\bibliography{new}
\bibliographystyle{icml2025}

%%%%%%%%%%%%%%%%%%%%%%%%%%%%%%%%%%%%%%%%%%%%%%%%%%%%%%%%%%%%%%%%%%%%%%%%%%%%%%%
%%%%%%%%%%%%%%%%%%%%%%%%%%%%%%%%%%%%%%%%%%%%%%%%%%%%%%%%%%%%%%%%%%%%%%%%%%%%%%%
% APPENDIX
%%%%%%%%%%%%%%%%%%%%%%%%%%%%%%%%%%%%%%%%%%%%%%%%%%%%%%%%%%%%%%%%%%%%%%%%%%%%%%%
%%%%%%%%%%%%%%%%%%%%%%%%%%%%%%%%%%%%%%%%%%%%%%%%%%%%%%%%%%%%%%%%%%%%%%%%%%%%%%%
\clearpage
\appendix
\onecolumn
\begin{figure}
    \centering
    \includegraphics[width=0.7\linewidth]{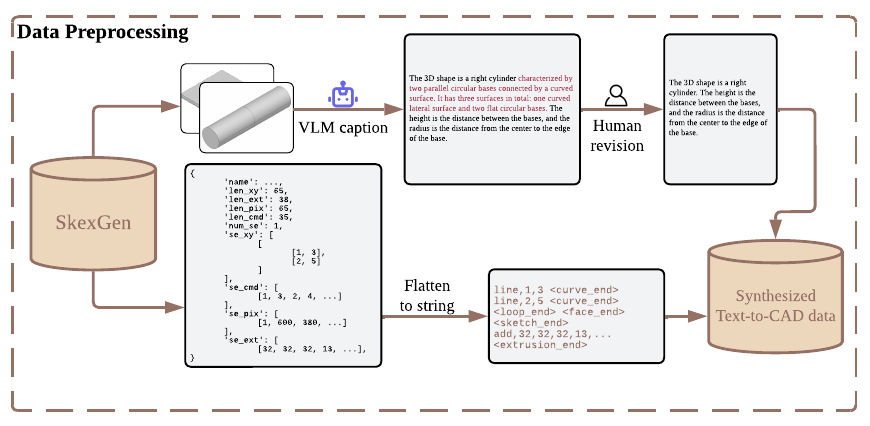}
    \caption{An overview of the data preprocessing steps. The original dataset is transformed into captions that serve as textual inputs, while the corresponding stringified CAD representations are used as ground truth references.}
    \label{fig:data-preprocessing}
\end{figure}

\begin{figure*}
    \centering
    \includegraphics[width=\linewidth]{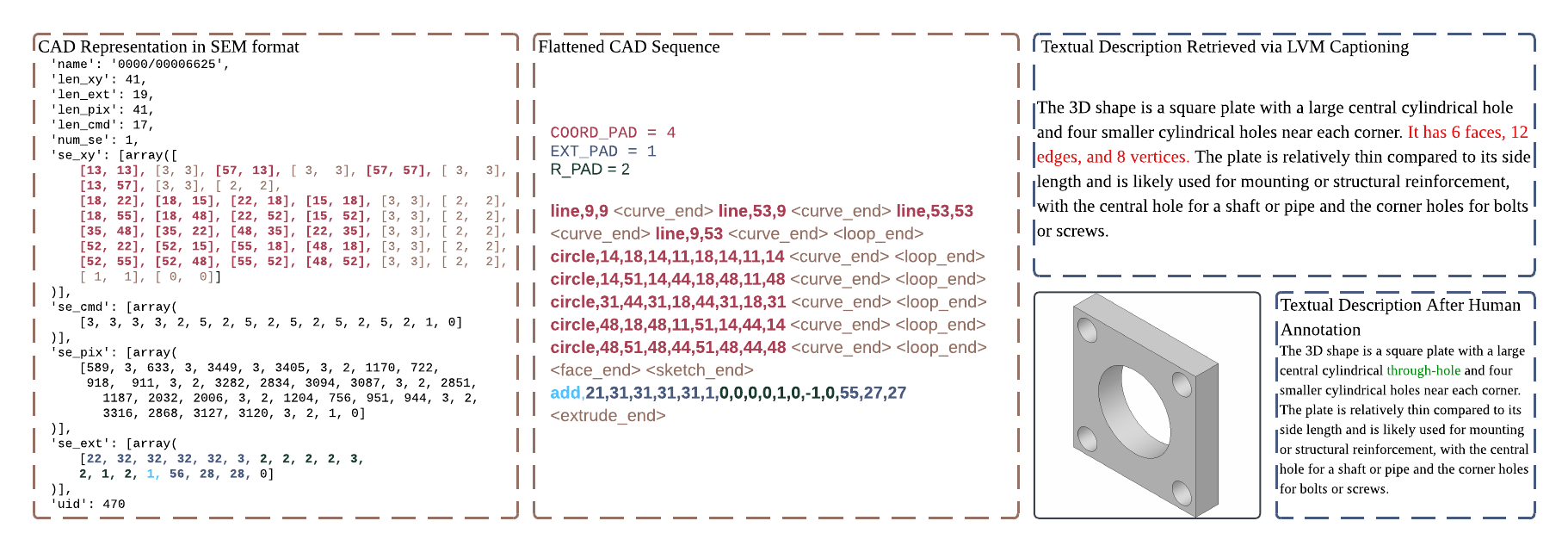}
    \caption{An overview of multiple CAD representations and their corresponding captions. Left: A CAD representation in the raw SEM format alongside its stringified sequence, with values highlighted in different colors based on the padding used for decoding. Right: Captions generated by the LVM and refined by human annotation. Phrases removed during human fine-tuning are marked in red, while those added by humans are marked in green. All representations and captions correspond to the same CAD figure, which is displayed in the bottom-right corner.}
    \label{fig:cad-rep}
\end{figure*}

\section{User Guidelines for Prompting}
A good prompt follows a structured description: (1) \textit{shape overview}, (2) \textit{shape details}, and (3) \textit{shape applications}. Given the varying shape complexity, we encourage but do not enforce describing item (2) and (3). Below is an example caption retrieved from Figure \ref{fig:additional-qualitative-1}, Row 2, Item 3, demonstrating this approach:

\begin{lstlisting}[caption={User guidelines for prompting.}, language=python]
"""
[Shape Overview] The 3D shape consists of a large, flat rectangular slab with two evenly 
spaced, identical cylindrical protrusions extending vertically from its surface. [Shape
Details (Optional)] The slab provides a stable base with significant length and width
compared to its thin height, while the cylinders are relatively short and have small
diameters. [Shape Applications (Optional)] The overall design is symmetrical and balanced, potentially serving as a mounting base or connector.
"""
\end{lstlisting}

\section{Additional Dataset Construction Detail} \label{apdx:additional-dataset-detail}

\subsection{Converting Raw Data into Strings} \label{apdx:string-output}

\paragraph{\model's String Format}
Our representation adopts the Sketch-and-Extrude Modeling (SEM) format, wherein a CAD instance is composed of sketches and extrusions. 
Each sketch is structured into multiple faces, and each face comprises multiple loops. 
Within each loop, geometric primitives such as lines, arcs, and circles are parameterized as follows:
\begin{itemize}
    \item \textbf{Line}: Represented by a line identifier and one coordinate.
    \item \textbf{Arc}: Defined by an arc identifier and two coordinates.\footnote{While lines and arcs generally require 2 and 3 coordinates for representation, respectively, this work leverages a simplified representation where the endpoints of lines and arcs are determined by the first point of the subsequent curve. If the loop is closed at the current curve, its endpoint is determined by the first curve in the loop.}
    \item \textbf{Circle}: Represented by a circle identifier and four coordinates.
\end{itemize}
Each extrusion is represented as a sequence formatted as \texttt{BVVTTTRRRRRRRRRSOO}, where the components are defined as follows:
\begin{itemize} 
    \item \texttt{B}: The boolean operation, selected from {\texttt{add}, \texttt{cut}, \texttt{intersect}}.
    \item \texttt{V}: The displacements of the top and bottom planes from the reference plane.
    \item \texttt{T}: The 3D translation vector.
    \item \texttt{R}: The 3D rotation, represented as a quaternion or equivalent.
    \item \texttt{S}: The scaling factor.
    \item \texttt{O}: The center of scaling.
\end{itemize}

\paragraph{Converting Source Data to \model's Format}
The original representation is derived from the SkexGen dataset \citep{xu_skexgen_2022}. 
Each CAD instance includes several components: sketch commands, sketch coordinates, and extrusion commands, which are stored in the \texttt{se\_cmd}, \texttt{se\_xy}, and \texttt{se\_ext} entries, respectively. 
The lengths of these entries correspond to the number of sketch-extrusion pairs within the complete CAD shape. 
To convert these entries into strings, we iteratively describe the sketches and extrusions in our format, ensuring that the resulting sequence reflects the chronological design order of the CAD process.

In iteration $i$, we select the $i$-th item from the \texttt{se\_xy}, \texttt{se\_cmd}, and \texttt{se\_ext} entries. 
For each digit in the \texttt{se\_cmd} array, we perform operations based on the command value as follows:
\begin{itemize}
\item \textbf{Command value $=5$}: Create a circle; use the first 4 items in \texttt{se\_xy} as XY coordinates and append the \texttt{<curve\_end>} token. Skip 5 positions in the \texttt{se\_xy} array.
\item \textbf{Command value $=4$}: Create an arc; use the first 2 items in \texttt{se\_xy} as XY coordinates and append the \texttt{<curve\_end>} token. Skip 3 positions in the \texttt{se\_xy} array.
\item \textbf{Command value $=3$}: Create a line; use the first item in \texttt{se\_xy} as an XY coordinate and append the \texttt{<curve\_end>} token. Skip 2 positions in the \texttt{se\_xy} array.
\item \textbf{Command value $=2$}: Mark the end of the loop by appending the \texttt{<loop\_end>} token. Skip 1 position in the \texttt{se\_xy} array.
\item \textbf{Command value $=1$}: Mark the end of the face by appending the \texttt{<face\_end>} token. Skip 1 position in the \texttt{se\_xy} array.
\item \textbf{Command value $=0$}: Mark the end of the sketch by appending the \texttt{<sketch\_end>} token. Skip 1 position in the \texttt{se\_xy} array.
\end{itemize}
Extrusions are represented by the 1D array \texttt{se\_ext}. The operation identifier is translated into a word, and the remaining values are flattened. To distinguish coordinates from special tokens, all coordinates are initially padded; they are subsequently unpadded based on the original padding values.
Figure \ref{fig:cad-rep} illustrates the conversion process from the SkexGen representation format to our stringified sequence.

\subsection{Generating Textual Instructions} \label{apdx:textual-input}
Textual instructions are generated in two steps: first, by applying LVM captioning on single-view images of CAD models; second, through human refinement of the generated captions to ensure clarity and accuracy.

Given a sequence representation, the CAD instance is rendered into an image, and captions are generated using GPT-4o. 
This step is designed to extract geometric properties, including the number of shapes, their dimensions, spatial arrangements, and other relevant details. 
The prompt used for this step is provided in Listing 1.

\begin{lstlisting}[caption={Prompts that are used for making captions. The first prompt is used to generate question-answer pairs, and the second prompt collects and summarizes the informations in the first prompt to yield the final caption.}, language=python]
{
    "Prompt1": "Propose a series of questions about the 3D shape and give the answers. The first question should ask for a detailed description and others should focus on the specific geometric properties, number, size proportions and positional relationship, and other details.",
    "Prompt2": "Based on the dialogue, please give a final description of the 3D shape. No more than 70 words."
}
\end{lstlisting}

The LVM-generated captions are further refined by human annotators to produce fine-grained captions that can serve as precise textual instructions. 
The human annotators follow these guidelines during the editing process:
\begin{itemize}
    \item \textbf{Ensuring Correspondence}: The description must accurately reflect the figure without any discrepancies.
    \item \textbf{Ensuring Succinctness}: The description should be as concise as possible while maintaining clarity and completeness.
    \item \textbf{Permission for Removal}: Figures that are excessively complex or challenging to describe may be excluded from the dataset. In practice, the annotators are permitted to mark the revised descriptions of such instances as "null". 
\end{itemize}
Figure \ref{fig:cad-rep} illustrates an example of how an image is captioned by the LVM and subsequently refined by human annotators.

\subsection{Dataset Construction}\label{apdx:dataset-construction}
The dataset construction process is illustrated in Figure \ref{fig:data-preprocessing}. 
Starting with a CAD representation from the original dataset, we generate a paired textual instruction and a stringified CAD representation. 
The textual instruction is created through the captioning process detailed in Section \ref{apdx:textual-input}, while the ground truth reference is obtained by converting the CAD formatting as described in Appendix \ref{apdx:string-output}.

\section{Additional Training Detail}

\begin{figure*}[t]
    \centering
    \includegraphics[width=0.9\linewidth]{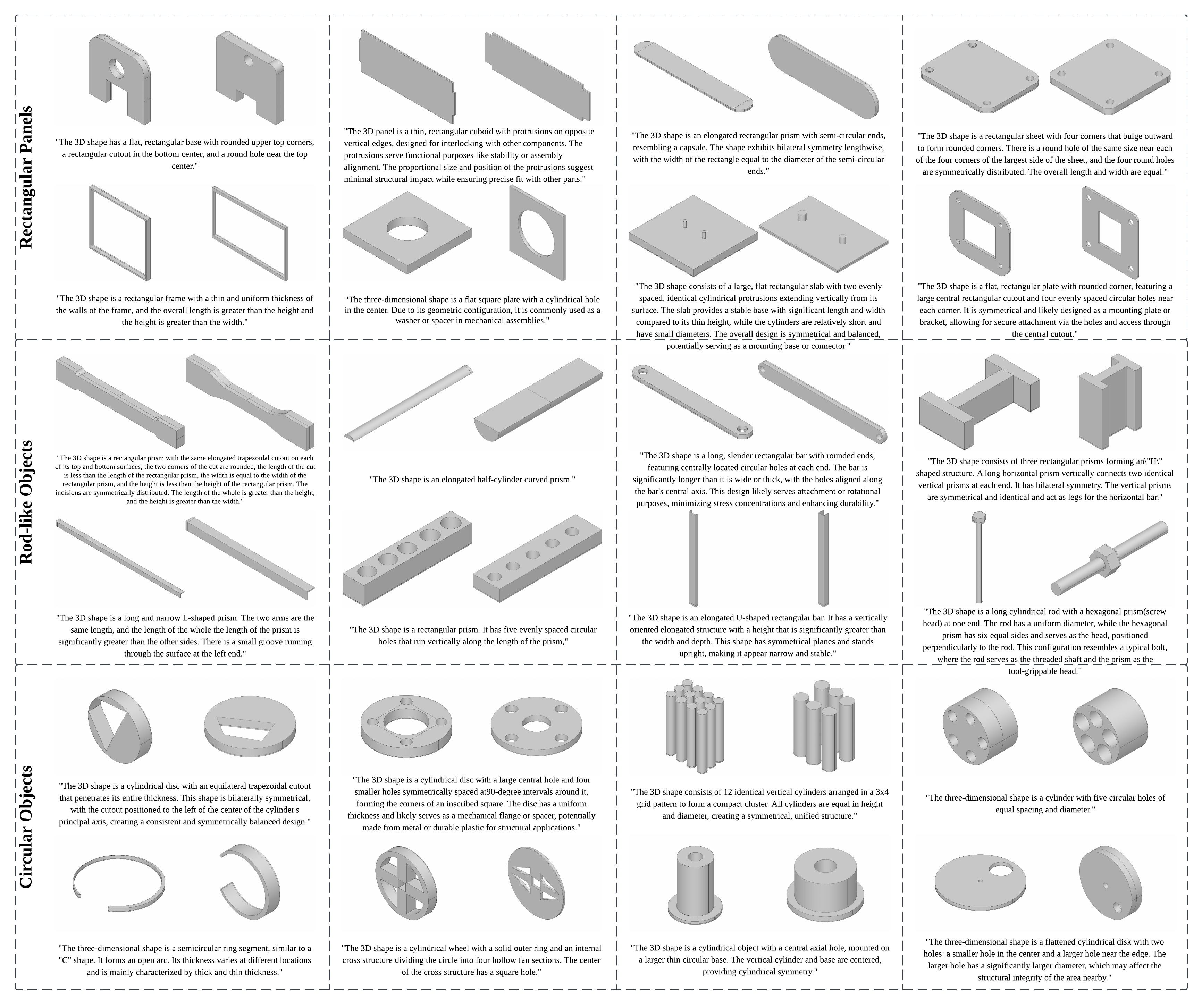}
    \caption{Additional qualitative results, Part 1. The results are grouped by categories such as panels and circular objects. In each sub-figure, the left image shows the figure rendered from the ground truth, while the right image displays the generation by \model. The corresponding textual instructions are provided at the bottom.}
    \label{fig:additional-qualitative-1}
\end{figure*}

\begin{figure*}[t]
    \centering
    \includegraphics[width=0.9\linewidth]{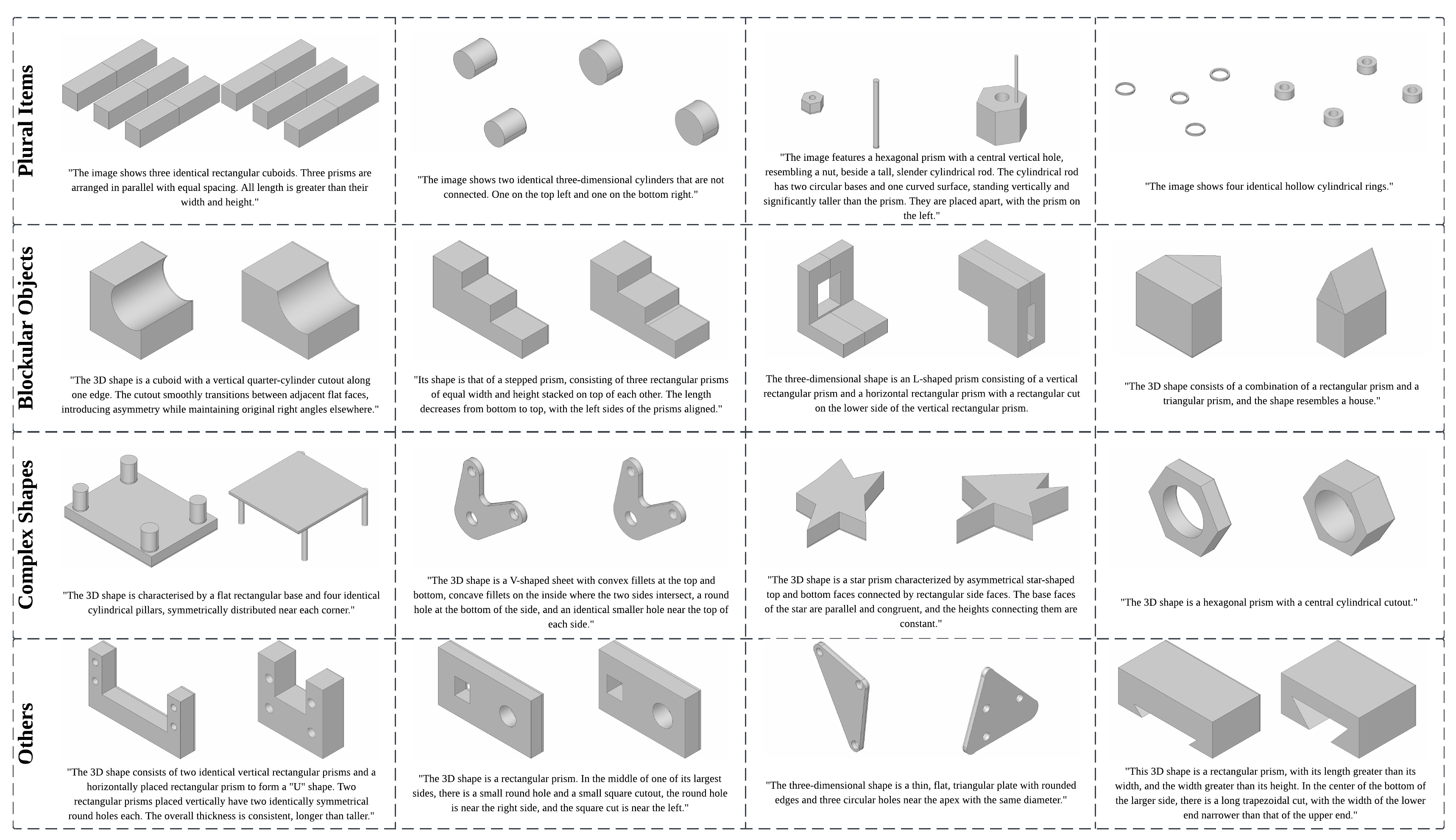}
    \caption{Additional qualitative results, Part 2. The results are grouped by categories such as multiple distinct items and complex shapes. In each sub-figure, the left image shows the figure rendered from the ground truth, while the right image displays the generation by \model. The corresponding textual instructions are provided at the bottom.}
    \label{fig:additional-qualitative-2}
\end{figure*}

\subsection{Sequential Learning}
We fine-tune a \texttt{LLaMA-3-8b-Instruct} by 40 epochs on 4 NVIDIA A6000-48GB SMX GPUs with a LoRA with rank 32. 
Further details regarding the fine-tuning process are provided in the Experiment Section of the main paper. 
The specific prompt used for the learning is as follows:
\begin{lstlisting}[caption={Prompt used for sequential learning. {description} refers to the actual textual commands of samples}, language=python]
    "Below is a description of a 3D shape:\n 
    {description}\n  
    Generate a Computer-Aided Design (CAD) command sequence of the 3D shape:\n" 
\end{lstlisting}

\subsection{Visual Feedback}
The visual feedback is collected as outlined in the main paper. 
The \texttt{llava-onevision-qwen2-7b-ov-chat} model is utilized to generate visual descriptions. 
For each input sequence produced by the post-SL \model, the corresponding rendered figure is evaluated by the model, which assigns a score ranging from 0 to 10. 
The prompt used for this scoring process is detailed in Listing 3:
\begin{lstlisting}[caption={Prompt used by LLaVA-OV for scoring an input figure. {description} refers to the textual of the sample.}, language=python]
    "You are a harsh grader for new CAD  designers' works. The following is a text description of a CAD figure that they designed and an image of a CAD instance. \n
    Description: {description} \n 
    Comment on this work for \n
    1. If the overall shape remains correct; \n 
    2. If the number of components are correct, especially the circular holes; \n 
    3. If the distribution of the components are natural, i.e. they are not clustered together or collide with each other. \n
    After that, give a score out of 10. Do not comment on issues such as texture, smoothness and colors."
\end{lstlisting}
The DPO procedure is conducted on 4 NVIDIA A6000-48GB SMX GPUs with a LoRA with rank 32. 
The training involves five iterative DPO/SFT rounds, which require approximately 2.5 days to complete.

\section{Additional Experimental Results}
\subsection{LVM Evaluation Setups} \label{apdx:additional-exp-setup}
As mentioned in the main paper, we used a \texttt{GPT-4o} model as the LVM evaluator. 
It is selected over \texttt{LLaMA-ov} because we attempt to prevent the impact of AI bias that makes it prefer its own generation \footnote{We acknowledge that this choice may introduce bias in the GPT-4o results. However, we have decided to proceed with it for two reasons: 1) our primary focus is on comparing Text2CAD with our model, and 2) the GPT-based generations during our experiment showed a big margin in LVM scores compared to other methods, so the impact of this bias is minimal.}.
The prompt we used for LVM evaluation is detailed in Listing 4:
\begin{lstlisting}[caption={Prompt used by GPT-4o for evaluation.}, language=python]
"
The following is a text description of a 3D CAD figure and an image of a CAD instance. Measure if the figure corresponds to the given description, and give a score in the scale of 10. Do not comment on issues such as texture, smoothness and colors \n description: {description}\n
"
\end{lstlisting}

\subsection{Human Evaluation Setups} \label{apdx:human-eval}
We generate a quadruple of outputs for each test set instruction. 
Each quadruple presents four rendered generations from \model$_{\text{SL}}$, \model, GPT-4o and Text2CAD, respectively.
The generations are tested by their correspondence between CAD shapes and instructions.
Six human judges (with college-level or higher education records) are asked to rank the generations\footnote{Due to the lack of overlapping, we obtain approximately 50 unique samples.} with the first place being the best model. The ranks are then collected and averaged to be the scores we presented as human evaluation.

\subsection{GPT Baselines}
The prompt we used for the GPT-4o baseline is detailed in Listing 5. 
\begin{lstlisting}[caption={Prompt used by GPT-4o for baseline comparison.}, language=python]
"   
    Below is a description of a 3D shape: 
    {description}
    Generate a Computer-Aided Design (CAD) command sequence of the 3D shape. The command sequence involves sketches such as lines, arcs, and circles, each marked by the endpoints, and extrusions that make the sketch into 3D volumes.

    Here are some examples:
    1. <few shot example>
    2. <few shot example>
    3. <few shot example>
    4. <few shot example>
    5. <few shot example>
    6. <few shot example>
    7. <few shot example>
    8. <few shot example>

    Now it's your turn. Remind that this is your description: {description}. No explanation is needed. Only return your final sequence, and in one line.
"
\end{lstlisting}

Alongside the current 8-shot version, we also tested a 3-shot GPT-4o model to reduce computation costs. 
However, the 3-shot model resulted in approximately a 92\% invalidity ratio, and the 8\% of renderable outputs were barely recognizable in relation to the prompt.
Given these issues, we have decided to use the 8-shot version as our baseline for GPT.

\subsection{Additional Statements on Text2CAD Results} \label{apdx:text2cad}
In the quantitative experiments, our setups are not fully aligned with those of Text2CAD. 
This discrepancy arises because when we used our test set prompts as input, we observed a performance degradation and a significant gap between our computed results and those reported by the authors.

We discovered that the discrepancy stems from the model's sensitivity to the level of detail in the prompt. 
Text2CAD performs well only with expert-level prompts, which contain step-by-step sketching guidelines. 
Our prompts, however, do not include this level of detail \footnote{We are concerned that the impact of detailed prompts containing step-by-step instructions and point coordinates is limited, as they may not be feasible in real-life scenarios.}.
To ensure consistency, in Table \ref{tab:quantative}, we report Text2CAD’s performance based on their expert prompts when computing the metrics they introduced. 
Specifically, for each item in the test set, \model\ and GPT-4o’s predictions were generated using our prompts, while Text2CAD’s predictions were generated using the expert prompt for the same item from their prompt base.

This approach aligns the results we reproduced with the reported scores from the original paper. 
To present a comprehensive and accurate study, we also report Text2CAD's results using our prompts and intermediate-level prompts in Table \ref{tab:text2cad}. 
The last two rows, \model\ and Text2CAD-our-prompt, are aligned as the same prompt is used.

By changing the prompt from the expert-level prompt in their database to an intermediate-level prompt, we observe a similar performance drop. 
This indicates that our prompting method does not degrade Text2CAD’s performance. 
Instead, it is an limitation stemmed from Text2CAD itself. 
Our model, using a simplified prompt, outperforms Text2CAD-expert. 
Given that the expert-level prompt from Text2CAD is too long and too specific to be feasible in the real designing process, we believe that our quantitative advantage over it is non-trivial.

\begin{table*}[h!]
    \centering
    \begin{tabular}{lccccc}
    \Xhline{2\arrayrulewidth}
    \multirow{2}{*}{\textbf{Method}} & \multicolumn{4}{c}{\textbf{F1}$\uparrow$} & \multirow{2}{*}{\textbf{CD$\downarrow$}} \\
    & Line & Arc & Circle  & Extrusion & \\
    \hline
        \textbf{Text2CAD-intermediate} & 66.65 & 4.85 & 47.62 & \textbf{93.56} & 146.15 \\ % our test set, t2c intermediate prompt 
        \textbf{Text2CAD-expert} & 79.59 & 42.79 & 69.45 & {92.13} & 30.23 \\ % our test set, t2c expert prompt
        \textbf{Text2CAD-our-prompt} & 54.42 & 0.92 & 18.42 & 75.37 & 235.91 \\ % our test set, our prompt. Also, invalidity = 24.16 according to our renderer
        \textbf{\model} & \textbf{83.71 }& \textbf{81.99} & \textbf{89.97} & {92.79} & \textbf{19.89} \\  
        \Xhline{2\arrayrulewidth}
    \end{tabular}
    \caption{Results of our model and different Text2CAD prompts on metrics Proposed by \citet{khan_text2cad_2024}. The suffix indicates the prompt type used for testing. Text2CAD-ours and \model\ are the most aligned pairs, while Text2CAD-expert and \model\ are the ones reported in the main paper.}
    \label{tab:text2cad}
\end{table*}

\subsection{Additional Quantative Results}
We report additional quantitative results in this section. 

\noindent \textbf{On Dataset Size.} 
The dataset used in our experiments is a subset of SkexGen \citep{xu_skexgen_2022}. 
Since human annotation is not scalable, we evaluate the trade-off between scalability and data quality. 
One such evaluation, discussed in the Ablation Study (Section \ref{sec:ablation}), demonstrates that, given the same number of training samples, data quality outweighs dataset scalability in terms of model performance.

Additionally, we investigate whether increasing dataset size can mitigate quality limitations by conducting an experiment on the full SkexGen-based \ttc\ dataset (~170k samples). 
The results, presented in Table \ref{tab:sequential-learning-data}, indicate that increasing dataset size does not significantly improve the visual quality of model generations. 
While a slight performance gain is observed with additional training samples, the improvement is marginal, and none of the w/o HA variants outperform the human-annotated counterpart.

\noindent \textbf{On Human Evaluation.} 
We conducted human evaluations across four models: GPT-4o, Text2CAD, \model, and \model\ trained only with the Sequential Learning stage. 
However, only the first three models are reported in Table \ref{tab:quantative}. The complete results of human evaluation are presented in Table \ref{tab:human-eval}.
As indicated by the evaluation, the two \model\ variants are preferred over the baselines, with the version incorporating Visual Feedback receiving higher rankings from human judges. 
This highlights the effectiveness of visual feedback in improving model performance.

\begin{figure}
    \begin{minipage}[b]{.49\linewidth}
    \centering
    \begin{tabular}{lcc}
    \Xhline{2\arrayrulewidth}
    \textbf{Method} & \textbf{LVM Score} $\uparrow$ & \textbf{IR} $\downarrow$ \\
    \hline
    % invalidity will be update again
         \model$_{\text{SL}}$ & 7.69 & 4.84\\
         \model$_{\text{SL}_{\text{w/o HA}\sim18k}}$ & 6.56 & 6.00 \\
         \model$_{\text{SL}_{\text{w/o HA}\sim170k}}$ & 6.60 & 9.04 \\
     \Xhline{2\arrayrulewidth}
    \end{tabular}
    \captionof{table}{LVM scores and invalidity ratios across different \model\ variants. All three models are trained using only the initial Sequential Learning stage. The suffix \texttt{w/o HA} indicates that the variant does not use human-annotated data, while the number denotes the size of the training set.}
    \label{tab:sequential-learning-data}
    
    \end{minipage}\hfill
    \begin{minipage}[b]{.45\linewidth}
        \centering
        \begin{tabular}{lc}
        \Xhline{2\arrayrulewidth}
        \textbf{Method} & \textbf{Avg. Rank} $\downarrow$ \\
        \hline
            \textbf{GPT-4o} \textit{-8shot} & 3.22 \\
            \textbf{Text2CAD} & 2.97 \\
            \textbf{\model}-\textit{SFT only} & 2.03 \\
            \textbf{\model} & 1.86 \\
         \Xhline{2\arrayrulewidth}
        \end{tabular}
        \captionof{table}{Human Evaluation Results. Human annotators ranked the generations of different methods based on their quality, with a lower rank indicating higher human preference.}
        \label{tab:human-eval}
    \end{minipage}
\end{figure}

\subsection{Additional Qualitative Results} \label{apdx:qualitative}
In this section, we present additional qualitative results. 
Figures \ref{fig:additional-qualitative-1} and \ref{fig:additional-qualitative-2} display these results, organized by CAD shape properties such as panels and circular objects. 
These examples demonstrate that our model can efficiently handle a variety of CAD shapes with distinct instructions, such as holes and frames. 
Furthermore, the model performs well in generating multiple identical objects, as shown in the first row of Figure \ref{fig:additional-qualitative-2}, and can effectively generate more complex shapes, such as stars and V-shapes.

\subsection{Text to Multiple CAD Figures}\label{apdx:text-to-multiple-cad}
\begin{figure*}[t!]
    \centering
    \includegraphics[width=0.8\linewidth]{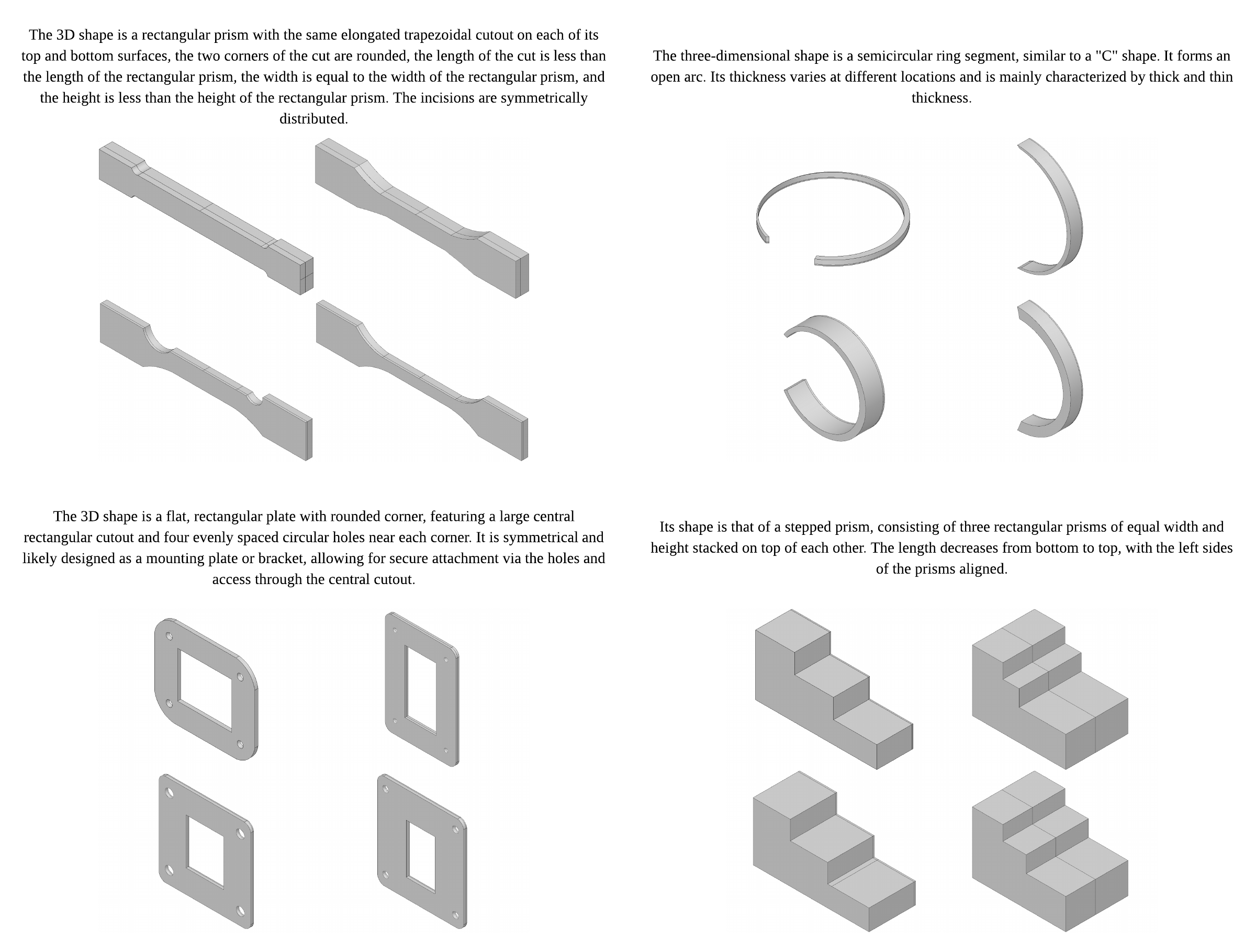}
    \caption{An overview of the generation of CAD instances with slight variations from a single prompt. In each sub-figure, the top-left image shows the ground truth generation, while the remaining three represent \model's outputs, which exhibit variations in thickness, width, and cutout size. The prompt is displayed at the top of each sub-figure.}
    \label{fig:multi-cad}
\end{figure*}

During inference, we set the temperature $t = 0.3$, top\_p $= 0.9$, and top\_k $= 50$ to enable non-deterministic generation. 
This configuration allows us to produce varied CAD figures that meet the instructed requirements, with slight differences between them. 
As a result, users can select the design that best aligns with their specific needs. 
Examples of such outputs are shown in Figure \ref{fig:multi-cad}. 
These results demonstrate that while adhering to the provided instructions, \model \ is capable of generating diversified outputs. 
The variations primarily affect attributes such as thickness, width, and the size of holes and cutouts, while maintaining the overall shape. 
This flexibility offers users a broader range of choices, thereby reducing the amount of additional work required when integrating such \ttc \ systems into industrial applications.

\begin{figure*}[t!]
    \centering
    \includegraphics[width=\linewidth]{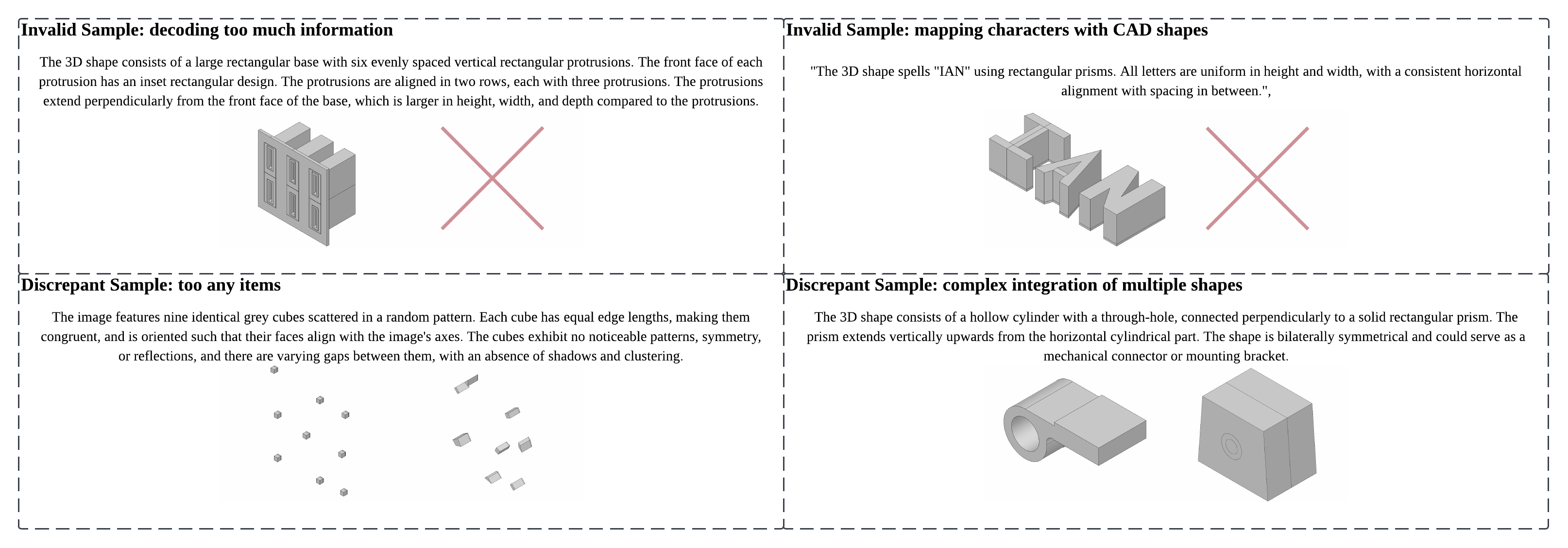}
    \caption{Invalid and discrepant samples. \model \ generates invalid samples when the instructions are too complex or involve word shape knowledge, and produces discrepant outcomes when there are too many distinct items to generate or when complex merges are required to form the final CAD instance.}
    \label{fig:failure}
\end{figure*}

\subsection{Failure Cases} \label{apdx:failure}
We identify two types of failures in our work: sequences that are not renderable and shapes that are rendered but misaligned with the intended design. 
We refer to the former as \textbf{Invalid Samples} and the latter as \textbf{Discrepant Samples}. 
Examples of both types of failures are shown in Figure \ref{fig:failure}.

In our analysis, samples are often invalid when the input instruction is too complex, meaning there are too many elements to be drawn. 
The case shown in the top-left corner of Figure \ref{fig:failure} involves more than 20 loops and 50 curves in the ground truth. 
Additionally, \model \ struggles to map CAD shapes to characters such as letters, resulting in failures when attempting to construct shapes that spell words or names.

Discrepancies between the rendered shapes and the intended design can occur when the input instruction involves too many distinct items. 
While \model \ demonstrates advanced capabilities in understanding numerical values compared to other models, handling more than 8 separate items remains a challenging task. 
In such cases, \model \ may either miscalculate the number of items to draw or generate incorrect shapes, as shown in the bottom-left corner of Figure \ref{fig:failure}. 
Furthermore, integrating multiple items into complex shapes is another frequent challenge for \model.

\section{Change Log}
\begin{enumerate}
    \item We updated our chamfer distance (CD) measurement results. The original outcome is computed between distributions and the new one is the accurate, pairwise result. This update does not affect the integrity of our work because the former was an upperbound of the latter measure and actually downgraded our model performance.
    \item We revised writing and updated the related work section according to the reviewers' recommendations.
\end{enumerate}

\end{document}